%% file: main.tex
\ifdefined\pdfminorversion\pdfminorversion=7\fi
\documentclass[sigconf,nonacm,balance=false]{acmart}

\AtBeginDocument{%
  }

\usepackage{stfloats}

\newcommand{\method}{MS-GPT}
\newcommand{\methodbase}{MS-GPT-Base}
\newcommand{\E}{\mathbb{E}}
\newcommand{\zstar}{z^{\star}}
\newcommand{\ind}{\mathbf{1}}

\begin{document}

\title[MS-GPT for MS/MS De Novo Structure Elucidation]{%
\texorpdfstring{MS-GPT: Rethinking MS/MS De Novo Structure Elucidation\\
as Spectrum-Induced Posterior Querying\\
of a Molecule-Language Model}%
{MS-GPT: Rethinking MS/MS De Novo Structure Elucidation as Spectrum-Induced Posterior Querying of a Molecule-Language Model}}

\settopmatter{authorsperrow=4, printacmref=false, printfolios=true}

\author{Xin Zhao}
\email{zhaoxin0807@sjtu.edu.cn}
\affiliation{%
  \department{MoE Key Lab of Artificial Intelligence, AI Institute, School of Computer Science}
  \institution{Shanghai Jiao Tong University}
  \city{Shanghai}
  \country{China}}

\author{Yumin Liu}
\email{ymliu@sjtu.edu.cn}
\affiliation{%
  \department{Instrumental Analysis Center}
  \institution{Shanghai Jiao Tong University}
  \city{Shanghai}
  \country{China}}

\author{Zhuo Li}
\email{zhuo.li@vipl.ict.ac.cn}
\affiliation{%
  \institution{Independent Researcher}
  \city{Beijing}
  \country{China}}

\author{Weichu Zheng}
\email{sjtu\_zwc0518@sjtu.edu.cn}
\affiliation{%
  \department{MoE Key Lab of Artificial Intelligence, AI Institute, School of Computer Science}
  \institution{Shanghai Jiao Tong University}
  \city{Shanghai}
  \country{China}}

\author{Feng Zhu}
\email{fzchem@sjtu.edu.cn}
\affiliation{%
  \department{Frontiers Science Center for Transformative Molecules;
  Shanghai Key Laboratory for Molecular Engineering of Chiral Drugs;
  School of Chemistry and Chemical Engineering;
  Zhangjiang Institute for Advanced Study}
  \institution{Shanghai Jiao Tong University}
  \city{Shanghai}
  \country{China}}

\author{Xiaokang Yang}
\authornote{Corresponding authors.}
\email{xkyang@sjtu.edu.cn}
\affiliation{%
  \department{MoE Key Lab of Artificial Intelligence, AI Institute, School of Computer Science}
  \institution{Shanghai Jiao Tong University}
  \city{Shanghai}
  \country{China}}

\author{Yaohui Jin}
\authornotemark[1]
\email{jinyh@sjtu.edu.cn}
\affiliation{%
  \department{MoE Key Lab of Artificial Intelligence, AI Institute, School of Computer Science}
  \institution{Shanghai Jiao Tong University}
  \city{Shanghai}
  \country{China}}

\author{Yanyan Xu}
\authornotemark[1]
\email{yanyanxu@sjtu.edu.cn}
\affiliation{%
  \department{MoE Key Lab of Artificial Intelligence, AI Institute, School of Computer Science}
  \institution{Shanghai Jiao Tong University}
  \city{Shanghai}
  \country{China}}
\renewcommand{\shortauthors}{Zhao et al.}

\begin{abstract}
Molecular structure elucidation from tandem mass spectra (MS/MS) is a central inverse problem in analytical chemistry.
Most existing approaches to MS/MS identification remain tied to reference libraries or predefined candidate sets, whereas \textit{de novo} methods aim to generate structures directly from spectra.
A common \textit{de novo} route predicts a molecular fingerprint from the spectrum and then decodes structures from it, enabling decoder pretraining on large molecule-only corpora.
However, this paradigm creates a training--inference mismatch: the decoder is trained on oracle fingerprints computed from molecules, but at inference it is queried with a noisy spectrum-induced fingerprint posterior that is typically collapsed to a single thresholded fingerprint.
We introduce \method{}, which recasts fingerprint-mediated \textit{de novo} elucidation as spectrum-induced posterior querying of a conditional molecule-language model.
\method{} conditions a molecule-language model on fingerprints and formulas, then converts the spectrum-induced posterior into a band of fingerprint queries near the oracle-fingerprint manifold through active-bit density calibration.
Candidates sampled across this band are pooled and ranked by generation frequency and posterior similarity.
A lightweight LoRA adapter further mitigates domain-specific posterior bias while preserving the pretrained molecular prior.
On NPLIB1 and MassSpecGym, \method{} sets a new state of the art, reaching Top-1/Top-10 exact-match accuracy of $17.1\%/27.6\%$ and $10.2\%/13.2\%$, respectively, and candidate-pool scaling improves recovery with a near-linear increase in inference cost.
The source code and model checkpoints are available at \url{https://github.com/VIKI623/MS-GPT}.
\end{abstract}

%  CCS + KEYWORDS  (required for papers over two pages)



\maketitle

% Page 1 is reserved for the title block and front matter; the main text
% starts at the top of page 2.
\clearpage

\section{Introduction}

Molecular structure elucidation from tandem mass spectra (MS/MS) is a central inverse problem in analytical chemistry, with applications in drug discovery, natural product research, environmental monitoring, and clinical diagnostics~\cite{bittremieux2022spectral}.
Existing computational approaches match query spectra to reference libraries~\cite{horai2010massbank}, score database candidates through simulated fragmentation~\cite{ruttkies2016metfrag,wang2021cfmid4}, infer molecular fingerprints for candidate ranking~\cite{duhrkop2015csifingerid,goldman2023mist}, or retrieve structures through cross-modal alignment~\cite{kalia2025jestr,chen2026flare}.
Although effective, these methods remain tied to reference spectra or predefined candidate sets, limiting their reach to previously characterized regions of chemical space.
\textit{De novo} structure elucidation instead generates candidate structures directly from a query spectrum, offering a route into the dark chemical space that reference-dependent methods leave unexplored~\cite{stravs2022msnovelist,butler2023ms2mol,bushuiev2024massspecgym}.

A practical route to \textit{de novo} structure elucidation is the two-stage fingerprint-mediated paradigm.
A spectrum encoder first predicts a molecular fingerprint from the MS/MS input, and a conditional decoder then generates structures from the predicted fingerprint~\cite{stravs2022msnovelist}.
Because fingerprints can be computed from molecular structures alone, the decoder can be pretrained on large molecule-only corpora without annotated spectra~\cite{le2020neuraldecipher,ucak2023molforge}.
Recent work in this paradigm has largely emphasized more expressive decoders, including sequence, diffusion-like, and flow-based architectures~\cite{han2025msbart,bohde2025diffms,sun2026mbgen,bohde2026frigid,nie2026flowms}.

However, improving the decoder architecture alone leaves a central mismatch unresolved: the decoder is trained on discrete oracle fingerprints computed from molecules, but inference supplies a noisy spectrum-induced fingerprint posterior $\pi$.
Most pipelines collapse this posterior to a single thresholded fingerprint before molecule generation, replacing a distribution over fingerprint bits with one deterministic point query.
We instead treat the posterior $\pi$ not as a fingerprint to be thresholded once, but as a distribution from which decoder queries should be constructed.
This view reformulates fingerprint-mediated MS/MS \textit{de novo} elucidation as spectrum-induced posterior querying of a conditional molecule-language model: the model learns a spectrum-free map from fingerprint-formula queries to compatible molecular structures, while spectral evidence enters at inference through posterior-derived queries rather than a single point estimate.

Here, we introduce \method{} to realize this posterior-querying formulation.
\method{} builds on SAFE-GPT~\cite{noutahi2024safe}, a pretrained autoregressive generator for molecular SAFE token sequences, and adapts it to condition on fingerprints and formulas.
\method{} addresses this oracle-to-posterior gap along two complementary axes: how the posterior is converted into queries the model can answer, and how the model is adapted to answer domain-shifted posterior queries accurately.
Along the first axis, active-bit density band calibration and fixed-size group querying keep posterior-derived queries near the oracle-fingerprint manifold while spanning plausible posterior operating points.
Along the second axis, posterior-aligned adaptation selectively updates the query-reading pathway, aligning the model with posterior-derived active-bit co-occurrence patterns while preserving the pretrained molecular prior.

% Lead figure: declared near the end of the page-2 left-column material so it
% appears at the top of the right column; discussion remains in experiments.
\begin{figure}[t]
\centering
\includegraphics[width=0.92\linewidth]{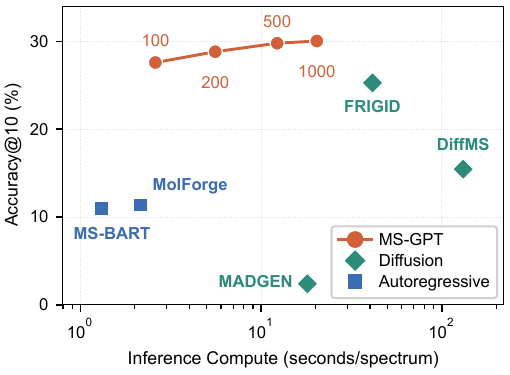}
\caption{NPLIB1 Top-10 exact-match accuracy versus per-spectrum inference compute (log scale). The \method{} curve spans candidate-pool sizes $B{=}100\text{--}1000$; other markers denote diffusion-like and autoregressive baselines. Candidate-pool scaling steadily improves recovery with a near-linear increase in inference compute, placing \method{} on a favorable accuracy--efficiency frontier.}
\Description{Scatter plot of per-spectrum inference compute on a logarithmic x-axis and Top-10 accuracy on the y-axis for NPLIB1. MS-GPT forms a curve annotated by candidate-pool size, while the diffusion-like and autoregressive baselines appear as single markers.}
\label{fig:efficiency}
\end{figure}
Our contributions are summarized as follows:
\begingroup
\makeatletter
% \@beginparpenalty=10000
\makeatother
\addtolength{\leftmargini}{-3.5pt}
\begin{itemize}
\item We reformulate fingerprint-mediated MS/MS \textit{de novo} elucidation as \textit{spectrum-induced posterior querying of a conditional molecule-language model}, moving beyond the common point-query approximation.
The conditional molecule-language model is pretrained on molecule-only data and conditioned on posterior-derived queries at inference.

\item We propose three components for spectrum-induced posterior querying.
\textit{Active-bit density band calibration} calibrates a query band near the oracle-fingerprint manifold.
\textit{Fixed-size group querying} distributes the candidate pool across posterior operating points in the band while holding its total size fixed.
\textit{Posterior-aligned adaptation} tunes query reading while preserving the pretrained molecular prior.

\item We establish new state-of-the-art exact-match accuracy on NPLIB1 and MassSpecGym.
Our ablation study shows that all three components provide complementary gains.
Candidate-pool scaling improves recall with a near-linear increase in inference cost.
\end{itemize}
\endgroup

\begin{figure*}[t]
\centering
\includegraphics[width=\textwidth]{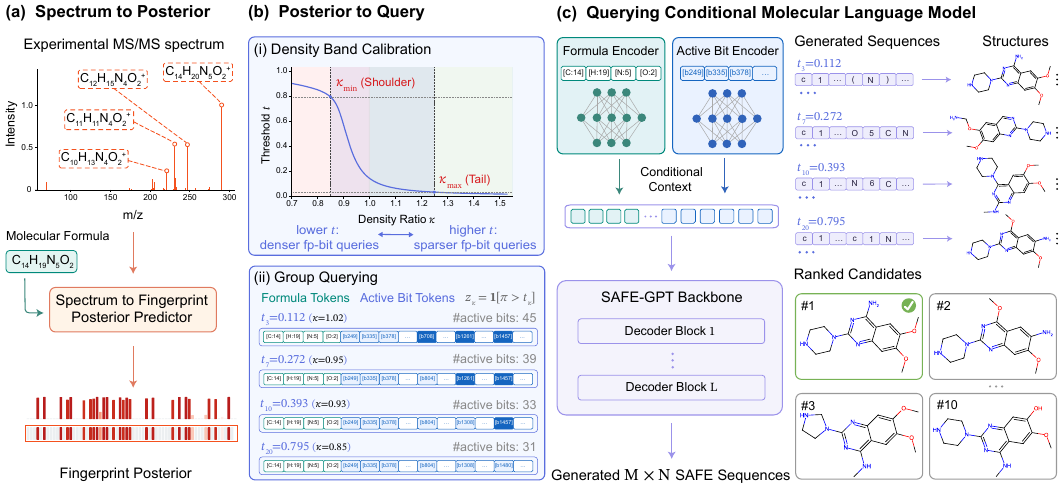}
\caption{MS-GPT inference as spectrum-induced posterior querying.
(a) A spectrum encoder maps an MS/MS spectrum and formula to a fingerprint posterior $\pi$.
(b) Active-bit density band calibration maps $\pi$ to a calibrated threshold interval, yielding multiple posterior-derived queries.
(c) The conditional molecule-language model generates SAFE sequences for each query; pooled candidates are grouped by InChIKey-14 and ranked by formula matching, generation frequency and posterior similarity.}
\Description{MS-GPT inference: spectrum-to-fingerprint posterior prediction, density band calibration, group querying, and conditional molecule-language model generation, followed by structure grouping, formula matching, and ranking.}
\label{fig:overview}
\end{figure*}

\section{Related Work}

\subsection{Molecule-Language Models}

Molecule-language models build on the serialization of molecular graphs as token sequences.
SMILES provides a compact line notation for chemical structures~\cite{weininger1988smiles}, while later representations such as SELFIES and SAFE improve validity, substructure locality, and sequence generation~\cite{krenn2020selfies,noutahi2024safe}.
Building on these notations, Transformer~\cite{vaswani2017attention} pretraining has produced transferable molecular representations for property prediction and representation learning, including SMILES Transformer, ChemBERTa, and MoLFormer~\cite{honda2019smilestransformer,chithrananda2020chemberta,ross2021molformer}.
Molecule strings have also supported cross-modal and generative models, including MolT5 for translating between molecules and text, GP-MoLFormer for scaled autoregressive generation, and SAFE-GPT for fragment-structured SAFE generation~\cite{edwards2022molt5,ross2024gpmolformer,noutahi2024safe}.
Together, these studies show that large-scale molecule-only corpora can endow sequence models with broad chemical priors.

Unlike prior molecule-language models for unconditional generation, property optimization, or text-conditioned design, our model conditionally maps molecular formulas and substructural fingerprints to compatible structures.
This conditioning formulation enables molecule-only pretraining for MS/MS elucidation, but it also creates an oracle-to-posterior gap: pretraining conditions on oracle fingerprints, whereas inference supplies a noisy spectrum-induced fingerprint posterior.
\method{} addresses this gap by making posterior-derived query construction the central mechanism linking spectral evidence to the pretrained molecular prior.

\subsection{Structure Elucidation from Mass Spectrometry}

Computational methods for MS/MS identification have traditionally relied on reference spectra or predefined candidate sets.
Spectral-library search compares query spectra with measured references~\cite{bittremieux2022spectral,horai2010massbank}; in silico fragmentation matches candidate-derived fragment masses to observed peaks~\cite{ruttkies2016metfrag}, while spectral prediction methods rank candidates by matching predicted spectra to the query~\cite{wang2021cfmid4,wang2025iceberg}; CSI:FingerID and MIST predict molecular fingerprints for candidate ranking~\cite{duhrkop2015csifingerid,goldman2023mist}; and learned spectral embeddings support exact-match and molecular analogue search~\cite{huber2021ms2deepscore,dejonge2023ms2query,bushuiev2025dreams}.
These approaches perform well when suitable references or candidates are available at inference time, but their search space is bounded by the chosen library or database.
\textit{De novo} structure elucidation relaxes this constraint by generating structures directly from spectra, thereby supporting the study of under-annotated natural products and the exploration of dark chemical space.

Some methods translate spectra into molecular strings~\cite{shrivastava2021massgenie,litsa2023spec2mol,butler2023ms2mol,neo2025fingerprints}.
MSNovelist instead popularized a fingerprint-mediated route from fingerprint prediction to SMILES decoding~\cite{stravs2022msnovelist}.
This route is attractive because its decoder can be pretrained on molecule-only corpora, as demonstrated by Neuraldecipher and MolForge~\cite{le2020neuraldecipher,ucak2023molforge}.
Recent work has broadened the decoder family: MS-BART adopts sequence-to-sequence generation~\cite{han2025msbart}; MADGEN combines scaffold retrieval with spectrum-guided generation~\cite{wang2025madgen}; DiffMS, MBGen, and FRIGID develop graph or masked diffusion decoders~\cite{bohde2025diffms,sun2026mbgen,bohde2026frigid}; and flow-matching variants generate structures through iterative probability flows or formula-constrained molecular graphs~\cite{mqawass2026msflow,nie2026flowms}.
Orthogonal formulations explore test-time tuned language models and iterative optimization with forward spectral simulation~\cite{mismetti2025tttms,manjrekar2026foam}.

\looseness=-1
Across these lines of work, recent progress has largely centered on more expressive decoder architectures, while query construction from spectrum-induced fingerprint posteriors remains less explored.
Fingerprint-mediated pipelines commonly collapse each posterior to a single thresholded query, discarding per-bit confidences and alternative structural hypotheses.
\method{} applies active-bit density band calibration to form a band of posterior-derived queries near the oracle-fingerprint manifold, distributes a candidate pool of fixed size across the band, and ranks the pooled candidates by generation frequency and posterior similarity.
This posterior-query perspective complements stronger spectrum encoders and more expressive decoders: improved query construction can better align the spectrum-induced posterior with the conditional molecule-language model.

\clearpage

\section{Method}

\subsection{Problem Formulation}

Given a tandem mass spectrum $x$ and its molecular formula $C$, \textit{de novo} structure elucidation aims to generate the underlying molecular structure $y$ directly, without a reference library or a predefined candidate set. Following the fingerprint-mediated paradigm, we introduce a discrete fingerprint variable $z\in\{0,1\}^{d}$, represented as a $d$-bit Morgan fingerprint~\cite{morgan1965algorithm,rogers2010ecfp}:
\begin{equation}
p(y\mid x,C)
:=
\E_{z\sim q_\phi(z\mid x,C)}
\big[p_\theta(y\mid z,C)\big],
\label{eq:factorization}
\end{equation}
where $q_\phi$ is the spectrum-induced distribution over fingerprints and $p_\theta$ is a fingerprint-formula-conditioned decoder. For any molecule $m$, the corresponding oracle fingerprint is $\zstar_m=\mathrm{MorganFP}(m)$, allowing $p_\theta$ to be pretrained on molecule-only corpora without annotated spectra.

This decoupling creates a mismatch between pretraining and inference: the decoder is pretrained on oracle fingerprints, whereas inference uses thresholded predictions derived from a \emph{spectrum-induced fingerprint posterior} $\pi(x,C)\in[0,1]^d$, whose entries estimate bitwise activation probabilities (Figure~\ref{fig:overview}(a)). Existing pipelines apply a shared threshold $t$ to $\pi$, producing a single fingerprint $\hat z=\ind[\pi>t]$, and draw all candidates from $p_\theta(y\mid\hat z,C)$. This \emph{point-query approximation} replaces the expectation in Eq.~\eqref{eq:factorization} with one deterministic operating point, discarding the per-bit confidences and the alternative structural hypotheses they encode. \method{} instead separates posterior query construction into two complementary steps: active-bit density band calibration determines the query interval, while fixed-size group querying constructs $M$ posterior queries at interpolated thresholds within the band and draws $N$ samples per query. Across these queries, the conditional molecule-language model exploits learned fingerprint-bit co-occurrence to assemble fingerprint-encoded substructures into compatible molecular structures. In our implementation, a separate MIST spectrum encoder~\cite{goldman2023mist} predicts $\pi$ for each spectral dataset, and $p_\theta$ is a fingerprint-formula-conditioned adaptation of SAFE-GPT~\cite{noutahi2024safe} that generates molecular structures $y$ as SAFE token sequences.

\subsection{Conditional Molecule-Language Model}

The conditional molecule-language model realizes $p_\theta(y\mid z,C)$ by generating SAFE sequence representations of molecular structures compatible with a fingerprint-formula query (Figure~\ref{fig:overview}(c)). We pretrain the decoder to maximize the conditional likelihood of these sequences given oracle fingerprints and formulas,
\begin{equation}
\max_\theta\; \E_{m \sim \mathcal{D}}\big[\log p_\theta(y_m \mid \zstar_m, C_m)\big],
\label{eq:pretrain}
\end{equation}
where each molecule $m$ in the molecule-only corpus $\mathcal{D}$ provides a SAFE sequence $y_m$, an oracle fingerprint $\zstar_m$, and a formula $C_m$, all derived from its molecular structure. Because the model never observes spectra during pretraining, the decoder remains spectral-domain independent: a single molecule-only pretrained decoder can be shared unchanged across downstream spectral datasets, with spectral evidence entering only at query time through $\pi$.

\paragraph{Architecture.} The conditional model adapts the SAFE-GPT decoder for dual fingerprint-formula conditioning, warm-started from its pretrained weights. A fingerprint encoder embeds the active bits of the fingerprint query $z$ into a set of conditioning tokens, and a formula encoder embeds the per-element atom counts of $C$. Cross-attention modules interleaved through the transformer blocks then attend to this conditioning context. During pretraining, $z$ is the oracle fingerprint $\zstar$; at inference, $z$ is a posterior-derived thresholded query. SAFE is a fragment-based representation, so the decoder emits ring systems and functional groups as coherent units, aligning the generation target with the substructural semantics that the Morgan fingerprint encodes.

\begin{figure*}[t]
\centering
\includegraphics[width=\textwidth]{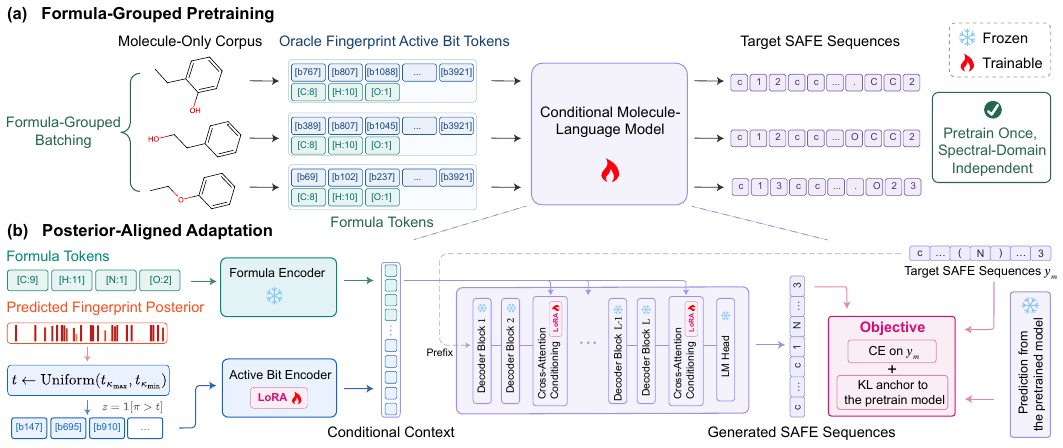}
\caption{
Pretraining and adaptation of the conditional molecule-language model.
(a) Molecule-only formula-grouped pretraining teaches the model to map oracle fingerprint-formula queries to SAFE sequences.
(b) Posterior-aligned adaptation trains lightweight LoRA modules on posterior-derived queries from the target spectral domain, while keeping the formula encoder, SAFE-GPT backbone, and language-model head frozen.
}
\Description{Two-panel schematic: (a) spectrum-free formula-grouped pretraining of the conditional molecule-language model on oracle fingerprint-formula queries; (b) posterior-aligned adaptation conditioning the model on thresholded spectrum-derived posterior queries with a cross-entropy and KL-anchor objective.}
\label{fig:pretrain-adapt}
\end{figure*}

\paragraph{Formula-grouped pretraining.} We group each mini-batch by exact molecular formula. Near-isomeric molecules that share a formula also share many fingerprint bits, placing structurally similar alternatives in direct contrast within a batch. This contrast encourages the model to resolve the remaining structural ambiguity under a fixed formula and oracle fingerprint, rather than exploiting that ambiguity as a generation shortcut (Figure~\ref{fig:pretrain-adapt}(a)).

\subsection{Posterior Query Construction}

\paragraph{Active-bit density band calibration.}
The conditional model is trained only on oracle fingerprints, so it implicitly expects queries from the oracle-fingerprint manifold. However, the oracle fingerprint is unavailable at inference time, and the fingerprint must instead be obtained by thresholding the spectrum-induced posterior:
\[
z(t)=\ind[\pi>t].
\]
The encoder's posterior $\pi$ is a separately learned estimate that carries prediction error, so a naively thresholded query can move away from the oracle-fingerprint manifold: too low a threshold admits many weak bits, whereas too high a threshold discards plausible substructures. An appropriate threshold cannot be chosen example by example at inference time, since no oracle fingerprint is available for comparison. We therefore calibrate at the population level, matching the active-bit density of thresholded posterior queries to the ground-truth active-bit density estimated from the encoder's training annotations.

Let $\mathcal{T}_{\mathrm{enc}}$ denote the spectrum encoder's training set and define the training-set active-bit density reference, measured by active-bit count, as
\begin{equation}
D_{\mathrm{enc}} =
\E_{(x,C,m) \sim \mathcal{T}_{\mathrm{enc}}}
\big[\|\mathrm{MorganFP}(m)\|_0\big],
\end{equation}
where $\|\cdot\|_0$ counts active bits in a binary fingerprint. Thus $D_{\mathrm{enc}}$ estimates the ground-truth fingerprint density on the same annotated spectra that train the encoder and shape the posterior $\pi$. For a threshold $t$, define
\begin{equation}
A(t) =
\E_{(x,C) \sim \mathcal{T}_{\mathrm{enc}}}
\big[\|\ind[\pi(x,C)>t]\|_0\big]
\end{equation}
as the expected active-bit count after thresholding the predicted posterior. Since $A(t)$ decreases with $t$, a target density ratio $\kappa$ determines the threshold
\begin{equation}
t_\kappa = A^{-1}(\kappa D_{\mathrm{enc}}),
\label{eq:opband}
\end{equation}
so that $A(t_\kappa)/D_{\mathrm{enc}}=\kappa$. Larger $\kappa$ lowers the threshold and retains additional lower-confidence bits, whereas smaller $\kappa$ raises the threshold and yields sparser queries.
An active-bit density band is thus a density-ratio interval $\mathcal{K}$ around the oracle-fingerprint active-bit density.
Through Eq.~\eqref{eq:opband}, this density band maps to a posterior-threshold interval that controls the mean active-bit count of the resulting queries over the encoder's training set.

\paragraph{Fixed-size group querying.} For candidate-pool size $B$, rather than drawing all candidates from a single thresholded query, we construct $M$ posterior queries from the threshold interval induced by the calibrated active-bit density band $\mathcal{K}=[\kappa_{\min},\kappa_{\max}]$ and draw $N$ samples from each, giving $B=M\times N$ (Figure~\ref{fig:overview}(b)). The distribution $q_\phi$ in Eq.~\eqref{eq:factorization} is induced by a threshold drawn uniformly from this interval and shared across all fingerprint bits. We invert Eq.~\eqref{eq:opband} only at the two band endpoints,
\[
t_{\kappa_{\max}} = A^{-1}(\kappa_{\max} D_{\mathrm{enc}}), \qquad
t_{\kappa_{\min}} = A^{-1}(\kappa_{\min} D_{\mathrm{enc}}),
\]
and linearly interpolate a prespecified number $M$ of thresholds between them,
\[
t_i = t_{\kappa_{\max}} + \frac{i-1}{M-1}\big(t_{\kappa_{\min}} - t_{\kappa_{\max}}\big), \quad i=1,\dots,M,
\]
thereby approximating
\begin{equation}
\begin{gathered}
p(y \mid x, C)
\approx \frac{1}{M}\sum_{i=1}^{M} p_\theta(y \mid z_i, C), \\
z_i = \ind[\pi > t_i].
\end{gathered}
\end{equation}
for $M>1$. The expectation in Eq.~\eqref{eq:factorization} can be approximated either by Monte Carlo sampling of the shared threshold or by deterministic quadrature. We use an equally spaced grid to cover the calibrated interval under a fixed query budget and avoid the additional variability introduced by random threshold sampling.
We choose $\mathcal{K}$ to span both sides of density match ($\kappa=1$), covering lower and higher active-bit densities relative to the oracle-fingerprint reference. Appendix~\ref{sec:kappa-diagnostics} presents post hoc diagnostic results for this choice, while Appendix~\ref{sec:band-config} provides the density-band configuration details. The point-query approximation adopted by prior fingerprint-mediated pipelines is recovered as the degenerate case $M{=}1$.

\subsection{Posterior-Aligned Adaptation}
\label{subsec:posterior-aligned-adaptation}

Density band calibration mitigates the aggregate mismatch in active-bit counts between posterior-derived queries and oracle fingerprints, but it does not resolve errors in which bits are active or how their dependencies differ from oracle fingerprints. Through molecule-only pretraining, the conditional model learns a molecular prior over fingerprint-bit co-occurrence: it can discount isolated active bits that conflict with the surrounding pattern and can still generate chemically coherent structures when some compatible substructures are missing from the query. However, if the spectrum encoder produces domain-specific posterior bias, thresholded queries may retain systematic bit-pattern shifts even inside the calibrated density band. These shifts affect how the pretrained decoder resolves competing chemically plausible completions. We therefore introduce a lightweight \emph{posterior-aligned adapter} that aligns the query-reading pathway with posterior-derived active-bit patterns in the target spectral domain, while preserving the pretrained molecular prior.

\paragraph{Adapter architecture.} Let $p_{\theta,\psi}$ denote the LoRA-adapted conditional model, where $\theta$ are the frozen pretrained weights and $\psi$ are trainable low-rank adapter parameters~\cite{hu2021lora}. We attach LoRA only to the query-reading pathway: the fingerprint encoder projections and the cross-attention modules that route fingerprint conditioning into the SAFE-GPT. The SAFE-GPT backbone, language-model head, and formula encoder remain frozen. This confines adaptation to how posterior-derived queries are encoded and routed into SAFE-token generation, rather than relearning the fingerprint-formula-conditioned molecular prior acquired during pretraining (Figure~\ref{fig:pretrain-adapt}(b)).

\paragraph{Adaptation training.} For each annotated pair $(x,m)$ in the domain training set $\mathcal{T}_{\mathrm{pa}}$, we retain the predicted posterior $\pi(x,C)$, the oracle fingerprint $\zstar_m$, the formula $C_m$, and the SAFE target sequence $y_m$. During adaptation, the decoder input is not the oracle fingerprint.
Instead, we sample one shared threshold $t$ uniformly from the same calibrated interval $[t_{\kappa_{\max}},t_{\kappa_{\min}}]$ used for query construction and form the thresholded query
\[
z_t(x,C)=\ind[\pi(x,C)>t].
\]
The adapter is trained with teacher-forced SAFE likelihood on these posterior-derived queries, plus a frozen-teacher KL anchor:
\begin{equation}
\begin{aligned}
\min_{\psi}\;
\E_{\substack{(x,C,m)\sim \mathcal{T}_{\mathrm{pa}}\\
t \sim \mathrm{Uniform}([t_{\kappa_{\max}},t_{\kappa_{\min}}])}}
\Big[
&\alpha_{g}\,
\mathcal{L}_{\mathrm{CE}}\big(p_{\theta,\psi}; y_m,z_t,C_m\big) \\
&+\lambda_{g}\,
\mathcal{L}_{\mathrm{KL}}\big(p_{\theta},p_{\theta,\psi}; y_m,z_t,C_m\big)
\Big],
\end{aligned}
\label{eq:pa_lora}
\end{equation}
where $\mathcal{L}_{\mathrm{CE}}$ is the next-token cross-entropy of $y_m$, and $\mathcal{L}_{\mathrm{KL}}$ is a token-level KL from the frozen model to the adapted model on the same posterior-derived query. The weights $(\alpha_g,\lambda_g)$ depend on recoverability: the \textbf{frozen-hit} group, where the frozen model already generates the ground-truth molecular structure at the default candidate-pool size, receives no CE and a strong KL anchor; the \textbf{residual} group, where the molecular structure is missed despite adequate posterior--oracle agreement, receives greater weight on CE than on KL to adapt the model to residual posterior bias; and the \textbf{low-coverage} group, where the molecular structure is missed with inadequate posterior--oracle agreement, receives weak CE and a KL anchor. Appendix~\ref{sec:adapt-details} specifies the recoverability grouping criterion and loss coefficients per group.

\section{Experiments}

\subsection{Experimental Setup}

\paragraph{Datasets and benchmarks.}
We evaluate under the known-formula protocol on two public MS/MS \textit{de novo} benchmarks: \textbf{NPLIB1}~\cite{duhrkop2021canopus} contains natural-product spectra from GNPS, and its training and test molecules are structurally similar, making it an in-distribution benchmark; \textbf{MassSpecGym}~\cite{bushuiev2024massspecgym} follows an MCES-based structurally held-out split, measuring generalization to structurally distant molecules.

\paragraph{Evaluation metrics.}
Following prior work on known-formula MS/MS \textit{de novo} evaluation, we report Top-$k$ accuracy, Top-$k$ Tanimoto similarity, and Top-$k$ MCES distance over ranked generated candidates at $k\in\{1,10\}$. \textbf{Top-$k$ accuracy} measures whether any of the top-$k$ candidates shares the same 2D structure (the first 14 InChIKey characters) with the ground-truth molecule. \textbf{Top-$k$ Tanimoto similarity} measures the highest structural similarity to the ground-truth molecule among the top $k$ candidates. \textbf{Top-$k$ MCES distance} measures the smallest graph-edit distance to the ground-truth molecule among the top $k$ candidates. For ablations, \textbf{Top-$k$ \% Close Match} (\%CM) measures the percentage of spectra whose top-$k$ generated candidates include a structure with RDKit topological-fingerprint Tanimoto similarity $\ge 0.675$~\cite{bohde2026frigid}. \textbf{Recall} measures whether any generated candidate shares the same 2D structure with the ground-truth molecule.
\paragraph{Baselines.}
We compare against representative methods from the main decoder families for this task. Autoregressive sequence baselines include Spec2Mol~\cite{litsa2023spec2mol}, MIST+\allowbreak Neuraldecipher~\cite{le2020neuraldecipher}, MIST+\allowbreak MSNovelist~\cite{stravs2022msnovelist}, MS-BART~\cite{han2025msbart}, and MIST+\allowbreak MolForge~\cite{ucak2023molforge}. Diffusion-like baselines include DiffMS~\cite{bohde2025diffms} and FRIGID~\cite{bohde2026frigid}; the scaffold-based baseline is MADGEN~\cite{wang2025madgen}. We also include FOAM~\cite{manjrekar2026foam}, which combines DiffMS with PubChem retrieval.
\paragraph{Implementation.}
The MIST spectrum encoder trained on the corresponding dataset produces the fingerprint posterior $\pi$, and the lightweight LoRA adapter is trained on posterior-derived queries from that spectral domain. The underlying conditional molecule-language model is pretrained for two epochs on approximately $100$M molecular structures, after excluding connectivity-equivalent structures from the validation and test splits of both benchmarks. Unless stated otherwise, we generate $B{=}100$ candidates using $M{=}20$ query groups and $N{=}5$ samples per group, and evaluate them after unified postprocessing. Appendix~\ref{sec:scaling} reports additional results on pretraining data scaling, while Appendix~\ref{sec:exp-details} provides full implementation details.

% Main-text float anchor: Table 1 remains at the top of page 7.
\begin{table*}[t]
\centering
\caption{\textit{De novo} structural elucidation performance on NPLIB1~\cite{duhrkop2021canopus} and MassSpecGym~\cite{bushuiev2024massspecgym}, assuming a known chemical formula. The \textbf{best} result in each column is bolded, and methods are ordered by Top-1 accuracy within each benchmark. Baseline entries are taken from the respective papers, with $^\dagger$ indicating revised MIST+MolForge accuracies reported by FRIGID~\cite{bohde2026frigid}. \method{}, FRIGID, and MolForge use the official MIST 2M-aug on NPLIB1 and MIST 4M-aug on MassSpecGym. For MS-BART on MassSpecGym, we report accuracy based on 2D InChIKey matching from Appendix C of MS-BART~\cite{han2025msbart}.}
\label{tab:main}
\setlength{\tabcolsep}{6pt}
\renewcommand{\arraystretch}{1.06}
\begin{tabular}{lcccccc}
\toprule
& \multicolumn{3}{c}{Top-1} & \multicolumn{3}{c}{Top-10} \\
\cmidrule(lr){2-4}\cmidrule(lr){5-7}
Model & Accuracy (\%)\,$\uparrow$ & MCES\,$\downarrow$ & Tanimoto\,$\uparrow$ & Accuracy (\%)\,$\uparrow$ & MCES\,$\downarrow$ & Tanimoto\,$\uparrow$ \\
\midrule
\multicolumn{7}{c}{\textit{NPLIB1}} \\
\midrule
Spec2Mol~\cite{litsa2023spec2mol}          & 0.00  & 48.57 & 0.12 & 0.00  & 17.58 & 0.16 \\
MADGEN~\cite{wang2025madgen}               & 2.10  & 20.56 & 0.22 & 2.39  & 12.69 & 0.27 \\
MIST+Neuraldecipher~\cite{le2020neuraldecipher} & 2.32  & 10.26 & 0.35 & 6.11  & 8.53  & 0.43 \\
MIST+MSNovelist~\cite{stravs2022msnovelist} & 5.40  & 13.10 & 0.34 & 11.04 & 8.57  & 0.44 \\
MIST+MolForge$^\dagger$~\cite{ucak2023molforge} & 5.62 & 20.72 & 0.43 & 11.36 & 10.55 & 0.52 \\
MS-BART~\cite{han2025msbart}               & 7.45  & 9.66  & 0.44 & 10.99 & 8.31  & 0.51 \\
DiffMS~\cite{bohde2025diffms}              & 8.34  & 10.46 & 0.35 & 15.44 & 8.54  & 0.47 \\
FRIGID~\cite{bohde2026frigid}              & 14.16 & 7.16 & 0.51 & 25.27 & 5.79 & 0.58 \\
\textbf{\method{} (ours)}                 & \textbf{17.09} & \textbf{6.69} & \textbf{0.53} & \textbf{27.59} & \textbf{5.11} & \textbf{0.61} \\
\midrule
\multicolumn{7}{c}{\textit{MassSpecGym}} \\
\midrule
MIST+MSNovelist~\cite{stravs2022msnovelist} & 0.00  & 39.84 & 0.06 & 0.00  & 18.83 & 0.15 \\
Spec2Mol~\cite{litsa2023spec2mol}          & 0.00  & 36.78 & 0.12 & 0.00  & 36.02 & 0.16 \\
MIST+Neuraldecipher~\cite{le2020neuraldecipher} & 0.00  & 22.93 & 0.14 & 0.00  & 21.76 & 0.16 \\
MS-BART~\cite{han2025msbart}               & 1.26  & 16.47 & 0.23 & 1.28  & 15.12 & 0.28 \\
MADGEN~\cite{wang2025madgen}               & 1.31  & 27.47 & 0.20 & 1.54  & 16.84 & 0.26 \\
FOAM (DiffMS + PubChem)~\cite{manjrekar2026foam} & 1.50 & 12.21 & 0.35 & 10.28 & \textbf{6.06} & \textbf{0.53} \\
DiffMS~\cite{bohde2025diffms}              & 2.30  & 13.96 & 0.28 & 4.25  & 11.68 & 0.39 \\
MIST+MolForge$^\dagger$~\cite{ucak2023molforge} & 3.29 & 31.77 & 0.31 & 5.65 & 16.67 & 0.40 \\
FRIGID~\cite{bohde2026frigid}              & 9.05 & 12.60 & \textbf{0.39} & 12.23 & 11.41 & 0.43 \\
\textbf{\method{} (ours)}                 & \textbf{10.22} & \textbf{12.09} & \textbf{0.39} & \textbf{13.21} & 10.55 & 0.45 \\
\bottomrule
\end{tabular}
\end{table*}

\subsection{Main Results}

Table~\ref{tab:main} summarizes performance. \method{} achieves the highest exact-match accuracy on both benchmarks, reaching Top-1/Top-10 accuracies of $17.09\%$/$27.59\%$ on NPLIB1 and $10.22\%$/$13.21\%$ on MassSpecGym. Compared with FRIGID, the strongest exact-match baseline, \method{} improves Top-1/Top-10 by $2.93$/$2.32$ percentage points on NPLIB1 and $1.17$/$0.98$ percentage points on MassSpecGym. These gains hold on both the in-distribution NPLIB1 benchmark and the structurally distant MassSpecGym benchmark. \method{} shares the same conditional molecule-language model across benchmarks; only the MIST spectrum encoder and lightweight LoRA adapter are trained for each benchmark.
\method{} also achieves the lowest Top-1 and Top-10 MCES on NPLIB1 and the lowest Top-1 MCES on MassSpecGym, while FOAM achieves the lowest Top-10 MCES on MassSpecGym.

\subsection{Component Ablation}

Table~\ref{tab:ablation-band-nplib1} reports a component ablation of \method{} on NPLIB1. Removing posterior-aligned adaptation gives \methodbase{}. Relative to \methodbase{}, \method{} improves Top-1/Top-10 accuracy by $1.22/3.66$ points. The larger gain at Recall indicates that posterior-aligned adaptation primarily improves recovery within the candidate pool.
Starting from \methodbase{}, we then isolate the two query-construction components. Replacing active-bit density band calibration with randomly sampled queries outside the calibrated band substantially reduces accuracy. Collapsing fixed-size group querying to a single threshold gives three point-query controls: a density-matched operating point at $\kappa{=}1.00$ and two operating points whose thresholds are selected on the validation set to maximize $F_1$ and cosine similarity respectively. Even the strongest point-query control falls below \methodbase{} by $2.20/6.11$ Top-1/Top-10 points. The MassSpecGym counterpart in Appendix~\ref{sec:ablation-msg} shows the same pattern.
\begin{table}[tbp]
\centering
\caption{Component ablation of \method{} on NPLIB1. Paired values report Top-1/Top-10.}
\label{tab:ablation-band-nplib1}
\small
\setlength{\tabcolsep}{2pt}
\renewcommand{\arraystretch}{1.06}
\begin{tabular*}{\linewidth}{@{\extracolsep{\fill}}lcccc@{}}
\toprule
Variant & Accuracy\,$\uparrow$ & Tanimoto\,$\uparrow$ & \%CM\,$\uparrow$ & Recall\,$\uparrow$ \\
\midrule
\textbf{\method{}} & \textbf{17.09/27.59} & \textbf{0.53/0.61} & \textbf{53.60/63.49} & \textbf{30.65} \\
\midrule
\multicolumn{5}{@{}l}{w/o Posterior-Aligned Adaptation} \\
\quad \methodbase{} & 15.87/23.93 & 0.51/0.59 & 51.04/62.76 & 24.54 \\
\midrule
\multicolumn{5}{@{}l}{w/o Posterior-Aligned Adaptation, Density Band Calibration} \\
\quad Outer-band querying & 9.04/16.61 & 0.40/0.48 & 34.80/46.40 & 17.46 \\
\midrule
\multicolumn{5}{@{}l}{w/o Posterior-Aligned Adaptation, Group Querying} \\
\quad Density match ($\kappa{=}1.00$) & 11.60/17.09 & 0.43/0.49 & 43.83/53.60 & 17.58 \\
\quad $F_1$ ($t{=}0.414$) & 13.68/17.83 & 0.49/0.53 & 48.23/55.07 & 18.32 \\
\quad Cosine ($t{=}0.519$) & 13.43/17.34 & 0.48/0.53 & 48.23/54.58 & 17.83 \\
\bottomrule
\end{tabular*}
\end{table}

\subsection{Posterior-Fidelity Analysis}
\label{sec:posterior-fidelity}

\begin{figure}[t]
\centering
\includegraphics[width=0.92\linewidth]{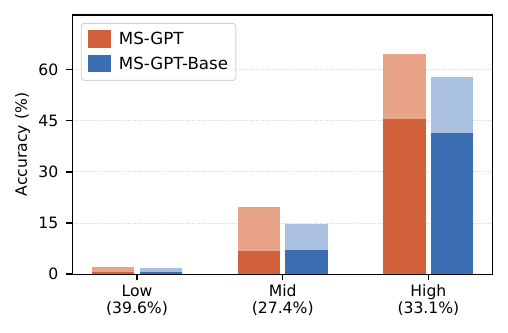}
\caption{NPLIB1 Top-1/Top-10 exact-match accuracy by posterior-fidelity bin. Parentheses indicate test-set proportions; dark segments show Top-1 accuracy and total bar heights show Top-10 accuracy.}
\Description{Grouped stacked bar chart of NPLIB1 accuracy by posterior fidelity. The Low, Mid, and High bins each contain bars for MS-GPT-Base and MS-GPT. Dark lower segments show Top-1 accuracy and pale upper segments extend each bar to Top-10 accuracy.}
\label{fig:strat-fidelity-nplib1}
\end{figure}

We stratify NPLIB1 by spectrum-induced posterior fidelity to investigate how much performance depends on posterior quality. For each test spectrum, posterior fidelity $T_c$ is the largest Tanimoto similarity between the oracle fingerprint and any posterior-derived query over the calibrated density band. This yields low ($T_c{<}0.5$), middle ($0.5{\le}T_c{<}0.7$), and high ($T_c{\ge}0.7$) bins~\cite{klein2026chemspace,franco2014fingerprint}. Figure~\ref{fig:strat-fidelity-nplib1} shows that \method{} accuracy rises sharply from the low-fidelity bin to the high-fidelity bin; the MassSpecGym counterpart in Appendix~\ref{sec:posterior-fidelity-msg} shows the same strong dependence. This benchmark-spanning dependence confirms that the fingerprint condition steers molecular generation: the model is not simply drawing plausible structures from the pretrained molecular prior independently of spectral evidence. Relative to \methodbase{}, posterior-aligned adaptation contributes most in the middle and high bins, especially at Top-10. The low-fidelity bin remains difficult for both models, identifying the spectrum-induced posterior as a central remaining bottleneck.

\subsection{Pretraining-Proximity Analysis}

\begin{figure}[t]
\centering
\includegraphics[width=0.92\linewidth]{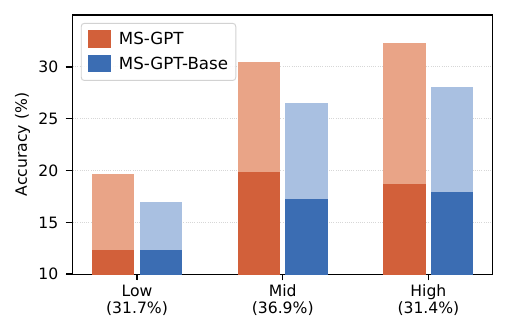}
\caption{NPLIB1 Top-1/Top-10 exact-match accuracy by pretraining-proximity bin. Parentheses indicate test-set proportions; dark segments show Top-1 accuracy and total bar heights show Top-10 accuracy.}
\Description{Grouped stacked bar chart of NPLIB1 accuracy by pretraining proximity. The Low, Mid, and High bins each contain bars for MS-GPT-Base and MS-GPT. Dark lower segments show Top-1 accuracy and pale upper segments extend each bar to Top-10 accuracy.}
\label{fig:strat-proximity-nplib1}
\end{figure}

We stratify NPLIB1 by pretraining proximity to investigate whether performance is driven mainly by close analogs in the fingerprint-formula pretraining set. Pretraining proximity is the maximum Tanimoto similarity between each test molecule and any pretraining molecule with the same formula, computed using $4096$-bit radius-2 Morgan fingerprints. The same cutoffs as in Section~\ref{sec:posterior-fidelity} define low, middle, and high proximity bins. Figure~\ref{fig:strat-proximity-nplib1} shows that accuracy improves from the low-proximity bin to the middle bin, suggesting that nearby structures provide a helpful conditional prior. However, from the middle to high bin, \method{} gains more modestly in Top-10 accuracy, while Top-1 accuracy decreases slightly. The gain of \method{} over \methodbase{} is not concentrated in the high-proximity bin: adaptation improves Top-10 recovery in all three bins and gives its largest Top-1 gain in the middle bin. These results suggest that nearby pretraining structures provide a useful prior, while close whole-molecule similarity alone does not explain the observed recovery pattern. This behavior is consistent with the molecule-language model learning reusable co-occurrence patterns between active-bit conditioning tokens and SAFE tokens. The corresponding MassSpecGym analysis in Appendix~\ref{sec:pretraining-proximity-msg} shows that accuracy increases across all three proximity bins, and indicates a stronger dependence on pretraining proximity on this benchmark, possibly reflecting its strict structural split.

% Main-text float anchors for the two stratified analyses.

\subsection{Inference Efficiency and Scaling}

\method{} generates each candidate in a single autoregressive pass, avoiding iterative graph generation or refinement. Figure~\ref{fig:efficiency} compares per-spectrum inference compute with Top-10 exact-match accuracy on NPLIB1. \method{} occupies a favorable accuracy--cost region: diffusion-like decoders require substantially more time, whereas autoregressive baselines are faster but substantially less accurate. Posterior querying therefore improves accuracy while retaining the efficiency of autoregressive molecular generation.

Autoregressive molecule generation also makes the candidate pool inexpensive to enlarge. In the candidate-pool scaling experiment, we keep the query grid fixed and draw more samples from each query, expanding the pooled candidate set with a near-linear cost increase. Figure~\ref{fig:efficiency} shows that Top-10 exact-match accuracy continues to improve as the candidate-pool size grows. The MassSpecGym counterpart in Appendix~\ref{sec:candidate-pool-scaling} shows the same trend, and Appendix~\ref{sec:candidate-pool-allocation} examines how a fixed candidate-pool size is allocated over query groups. Detailed case studies are provided in Appendix~\ref{sec:prediction-case-studies}.

\section{Conclusion}

We introduced \method{}, a spectrum-induced posterior-querying framework for fingerprint-mediated MS/MS \textit{de novo} structure elucidation. Instead of collapsing the fingerprint posterior to one point estimate, \method{} converts it into a calibrated band of posterior-derived queries near the oracle-fingerprint manifold and queries a conditional molecule-language model pretrained on a molecule-only corpus. Active-bit density band calibration, fixed-size group querying, and posterior-aligned adaptation bridge the oracle-to-posterior gap while preserving the pretrained molecular prior. On NPLIB1 and MassSpecGym, \method{} sets a new state of the art in exact-match accuracy; ablations show complementary gains from all three components. Candidate-pool scaling further improves Top-10 recovery with near-linear growth in inference compute. These results frame MS/MS \textit{de novo} elucidation as posterior querying over molecule-language models, with future progress depending on more faithful spectrum-induced posteriors and stronger conditional molecular generation.

%  LIMITATIONS AND ETHICAL CONSIDERATIONS  (REQUIRED by KDD AI4Sci CFP)
%  Must stay within the main content pages.
\section{Limitations and Ethical Considerations}

Our study is limited to two public MS/MS benchmarks under the known-formula setting and depends on a fixed-dimensional fingerprint posterior predicted by the spectrum encoder; missing or incorrect posterior information can therefore affect candidate generation.
The datasets contain no human participants or private data, so informed consent and institutional ethics review are not applicable. Practical applications, such as compound identification in drug discovery and environmental monitoring, require expert review and experimental validation of the predicted structures.

%  GENERATIVE AI USAGE  (REQUIRED by KDD AI4Sci CFP)
\section{Generative AI Usage}

Generative AI tools assisted with language polishing of the authors' original draft and code debugging. All suggested textual revisions and code changes were reviewed and verified by the authors.

%  End of the 8 main content pages. Force References onto a fresh page
%  (CFP: "8 content pages as main paper, followed by References").
\clearpage

\bibliographystyle{ACM-Reference-Format}
\bibliography{references}

%  APPENDIX  (no page limit; same format as the body)
\clearpage
\appendix
\raggedbottom

\section{Additional Analyses}

\subsection{Component Ablation}
\label{sec:ablation-msg}

Table~\ref{tab:ablation-band-msg} presents the MassSpecGym counterpart to the NPLIB1 ablation in Table~\ref{tab:ablation-band-nplib1}. \method{} leads consistently in exact-match accuracy, structural similarity, and close-match rate. Removing posterior-aligned adaptation causes a broad but moderate decline, indicating that adaptation helps the pretrained generator respond to posterior-derived conditions. Query construction has a stronger effect: sampling outside the calibrated density band weakens performance, while every single-threshold control falls short of grouped querying. The density-matched query is the strongest of these point controls, but it does not recover the benefit of covering multiple posterior conditions. This pattern mirrors NPLIB1 and shows that density calibration, group coverage, and posterior-aligned adaptation remain complementary in the more structurally shifted setting.

\begin{table}[H]
\centering
\caption{Component ablation of \method{} on MassSpecGym. Paired values report Top-1/Top-10.}
\label{tab:ablation-band-msg}
\small
\setlength{\tabcolsep}{2pt}
\renewcommand{\arraystretch}{1.06}
\begin{tabular*}{\linewidth}{@{\extracolsep{\fill}}lcccc@{}}
\toprule
Variant & Accuracy\,$\uparrow$ & Tanimoto\,$\uparrow$ & \%CM\,$\uparrow$ & Recall\,$\uparrow$ \\
\midrule
\textbf{\method{}} & \textbf{10.22/13.21} & \textbf{0.39/0.45} & \textbf{34.54/43.44} & \textbf{14.49} \\
\midrule
\multicolumn{5}{@{}l}{w/o Posterior-Aligned Adaptation} \\
\quad \methodbase{} & 9.27/12.75 & 0.38/0.44 & 32.97/42.61 & 13.23 \\
\midrule
\multicolumn{5}{@{}l}{w/o Posterior-Aligned Adaptation, Density Band Calibration} \\
\quad Outer-band querying & 3.74/5.79 & 0.28/0.33 & 22.78/29.60 & 5.96 \\
\midrule
\multicolumn{5}{@{}l}{w/o Posterior-Aligned Adaptation, Group Querying} \\
\quad Density match ($\kappa{=}1.00$) & 7.97/10.34 & 0.37/0.42 & 32.63/40.53 & 10.58 \\
\quad $F_1$ ($t{=}0.302$) & 7.42/9.69 & 0.36/0.41 & 31.62/38.69 & 9.87 \\
\quad Cosine ($t{=}0.424$) & 6.70/9.06 & 0.35/0.40 & 29.59/36.72 & 9.16 \\
\bottomrule
\end{tabular*}
\end{table}

\subsection{Posterior-Fidelity Analysis}
\label{sec:posterior-fidelity-msg}

\begin{figure}[!t]
\centering
\includegraphics[width=0.92\linewidth]{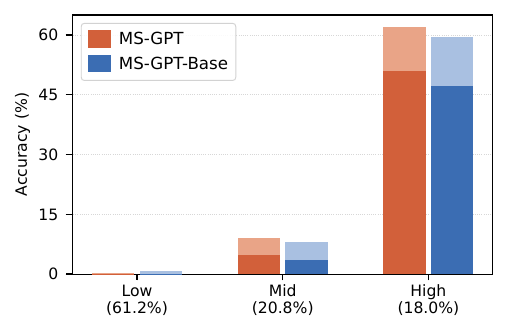}
\caption{MassSpecGym Top-1/Top-10 exact-match accuracy by posterior-fidelity bin. Parentheses indicate test-set proportions; dark segments show Top-1 accuracy and total bar heights show Top-10 accuracy.}
\Description{Grouped stacked bar chart of MassSpecGym accuracy by posterior fidelity. The Low, Mid, and High bins each contain bars for MS-GPT-Base and MS-GPT. Dark lower segments show Top-1 accuracy and pale upper segments extend each bar to Top-10 accuracy.}
\label{fig:strat-fidelity-msg}

\includegraphics[width=0.92\linewidth]{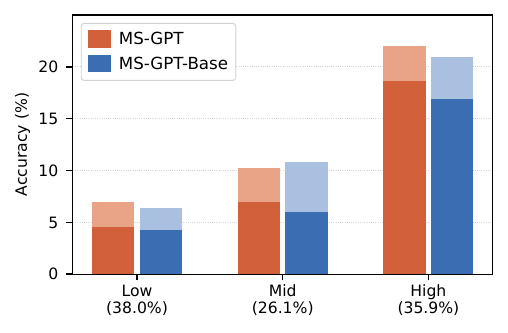}
\caption{MassSpecGym Top-1/Top-10 exact-match accuracy by pretraining-proximity bin. Parentheses indicate test-set proportions; dark segments show Top-1 accuracy and total bar heights show Top-10 accuracy.}
\Description{Grouped stacked bar chart of MassSpecGym accuracy by pretraining proximity. The Low, Mid, and High bins each contain bars for MS-GPT-Base and MS-GPT. Dark lower segments show Top-1 accuracy and pale upper segments extend each bar to Top-10 accuracy.}
\label{fig:strat-proximity-msg}
\end{figure}

Figure~\ref{fig:strat-fidelity-msg} extends the posterior-fidelity analysis to MassSpecGym. Accuracy separates sharply across the three bins: both variants recover almost no exact matches in the low-fidelity group, improve in the middle group, and perform substantially better in the high-fidelity group. The shared trend of \method{} and \methodbase{} shows that the quality of the spectrum-induced posterior remains a primary determinant of recovery under structural shift. The advantage of \method{} emerges mainly once the posterior provides informative structural evidence and is clearest in the high-fidelity group. Posterior-aligned adaptation therefore helps translate informative queries into broader candidate-pool recovery, but does not compensate for poorly grounded queries. The low-fidelity group remains the main target for improving the spectrum encoder and posterior construction.

% Keep the paired stratification figures together at the top of one column.

\subsection{Pretraining-Proximity Analysis}
\label{sec:pretraining-proximity-msg}

Figure~\ref{fig:strat-proximity-msg} extends the pretraining-proximity analysis to MassSpecGym. Both \method{} and \methodbase{} improve in Top-1 and Top-10 accuracy from the low-proximity group to the high-proximity group, with the largest increase between the middle and high groups. MassSpecGym is structurally held out, and connectivity-equivalent evaluation structures are additionally excluded from the corpus for fingerprint--formula conditional pretraining. A high-proximity neighbor is therefore structurally similar but cannot be the exact target. The increasing accuracy suggests that nearby pretraining structures provide a useful conditional prior. However, the model pretrained from scratch on the official MassSpecGym 50M-molecule corpus retains much of the main model's recovery performance under the strict structural split, as shown in Appendix~\ref{sec:alignment-50m-decoder}. Together, these results suggest that close pretraining analogs improve recovery, while transferable structural patterns support generation even in their absence.

% Queue two paired blocks for the tops of the next two columns.
\begin{figure}[!t]
\centering
\includegraphics[width=0.92\linewidth]{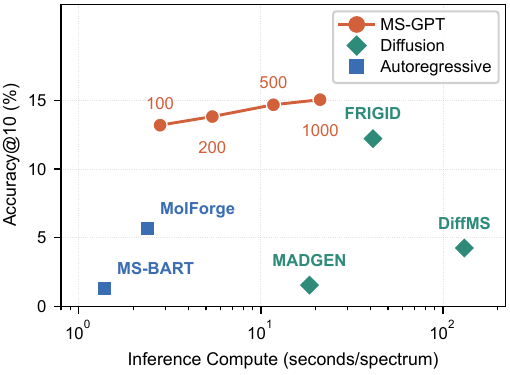}
\caption{MassSpecGym Top-10 exact-match accuracy versus per-spectrum inference compute (log scale). The \method{} curve spans candidate-pool sizes $B{=}100\text{--}1000$; other markers denote baseline methods.}
\Description{Scatter plot of per-spectrum inference compute on a logarithmic x-axis and Top-10 exact-match accuracy on the y-axis for MassSpecGym. MS-GPT forms a curve annotated by candidate-pool size, while baseline methods appear as single markers.}
\label{fig:efficiency-msg}

\begingroup
\makeatletter
\def\@captype{table}
\makeatother
\caption{Candidate-pool scaling of \method{} on NPLIB1 and MassSpecGym. Compute combines generation and Top-10 postprocessing, reported for an NVIDIA RTX 6000 Ada GPU with a 16-core Intel CPU.}
\label{tab:candidate-pool-scaling}
\setlength{\tabcolsep}{3pt}
\renewcommand{\arraystretch}{1.0}
\begin{tabular}{rcccc}
\toprule
$B$ & Compute (s/spec)\,$\downarrow$ & Top-1\,$\uparrow$ & Top-10\,$\uparrow$ & Recall\,$\uparrow$ \\
\midrule
\multicolumn{5}{c}{\textit{NPLIB1}} \\
\midrule
$100$ & $2.60$ & $17.09$ & $27.59$ & 30.65 \\
$200$ & $5.57$ & $16.61$ & $28.82$ & 32.23 \\
$500$ & $12.26$ & $16.85$ & $29.79$ & 35.53 \\
$1000$ & $20.32$ & $16.61$ & $30.04$ & 38.46 \\
\midrule
\multicolumn{5}{c}{\textit{MassSpecGym}} \\
\midrule
$100$ & $2.80$ & $10.22$ & $13.21$ & 14.49 \\
$200$ & $5.42$ & $10.40$ & $13.84$ & 15.67 \\
$500$ & $11.73$ & $10.67$ & $14.70$ & 17.44 \\
$1000$ & $21.16$ & $10.75$ & $15.07$ & 18.77 \\
\bottomrule
\end{tabular}
\endgroup

\begingroup
\makeatletter
\def\@captype{table}
\makeatother
\centering
\caption{Pretraining data scaling of \methodbase{} on NPLIB1. Rows list one-epoch checkpoints by cumulative structures seen. Paired values report Top-1/Top-10. Cumulative GPU-hours are estimated across eight NVIDIA A800 GPUs. Bold marks the best result.}
\label{tab:scaling}
\small
\setlength{\tabcolsep}{2pt}
\renewcommand{\arraystretch}{1.06}
\begin{tabular*}{\linewidth}{@{\extracolsep{\fill}}lccc@{}}
\toprule
Structures Seen & GPU Hours & Accuracy\,$\uparrow$ & Tanimoto\,$\uparrow$ \\
\midrule
$5$M    & $7.8$    & $9.16/18.19$ & $0.381/0.476$ \\
$10$M   & $14.9$   & $12.45/21.25$ & $0.428/0.520$ \\
$50$M   & $71.8$   & $14.41/24.54$ & $0.479/0.568$ \\
$100$M  & $143.1$  & $15.51/\mathbf{25.76}$ & $0.503/0.587$ \\
$500$M  & $712.0$  & $17.09/25.15$ & $0.518/0.591$ \\
$1$B    & $1422.5$ & $\mathbf{17.34}/25.27$ & $\mathbf{0.521/0.594}$ \\
\bottomrule
\end{tabular*}
\endgroup
\end{figure}

\begin{figure}[!t]
\centering
\includegraphics[width=0.92\linewidth]{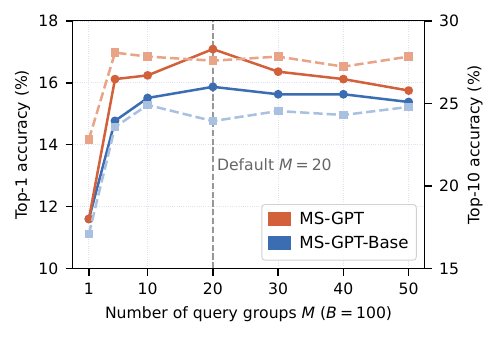}
\caption{NPLIB1 Top-1/Top-10 exact-match accuracy across query-group allocations at fixed candidate-pool size $B{=}100$. Increasing $M$ refines the grid over the calibrated density band while reducing samples per group ($N{\approx}100/M$). Solid circles and dashed squares denote Top-1 and Top-10 accuracy on the left and right axes respectively; the vertical line marks the default $M{=}20,N{=}5$.}
\Description{Dual-axis line plot of NPLIB1 accuracy across query-group allocations at fixed candidate-pool size 100. Orange MS-GPT and blue MS-GPT-Base curves pair solid circles with Top-1 and dashed squares with Top-10. A vertical dashed line marks the default 20 query groups and 5 samples per group.}
\label{fig:candidate-pool-allocation}

\includegraphics[width=0.92\linewidth]{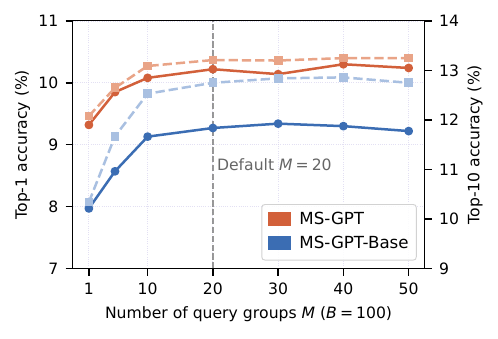}
\caption{MassSpecGym Top-1/Top-10 exact-match accuracy across query-group allocations at fixed candidate-pool size $B{=}100$. Increasing $M$ refines the grid over the calibrated density band while reducing samples per group ($N{\approx}100/M$). Solid circles and dashed squares denote Top-1 and Top-10 accuracy on the left and right axes respectively; the vertical line marks the default $M{=}20,N{=}5$.}
\Description{Dual-axis line plot of MassSpecGym accuracy across query-group allocations at fixed candidate-pool size 100. Orange MS-GPT and blue MS-GPT-Base curves pair solid circles with Top-1 and dashed squares with Top-10. A vertical dashed line marks the default 20 query groups and 5 samples per group.}
\label{fig:candidate-pool-allocation-msg}

\end{figure}

\subsection{Inference Efficiency and Scaling}
\label{sec:candidate-pool-scaling}

Figure~\ref{fig:efficiency-msg} plots MassSpecGym Top-10 exact-match accuracy against per-spectrum inference compute; Table~\ref{tab:candidate-pool-scaling} reports the corresponding fixed-grid sweeps for NPLIB1 and MassSpecGym. Across both sweeps, Top-10 accuracy and recall increase as the candidate pool grows, whereas Top-1 changes more modestly, with small fluctuations on NPLIB1. Together, these trends indicate that additional sampling broadens candidate coverage more reliably than it alters the leading prediction. Although the gains taper at larger candidate-pool sizes, \method{} still attains strong Top-10 exact-match accuracy on MassSpecGym at comparatively low per-spectrum inference compute.

\subsection{Allocation at Fixed Candidate-Pool Size}
\label{sec:candidate-pool-allocation}

Figures~\ref{fig:candidate-pool-allocation} and~\ref{fig:candidate-pool-allocation-msg} examine how a fixed candidate-pool size is divided between query diversity and per-query sampling depth. The $M=1$ control uses the density-matched point query; larger grids interpolate across the calibrated band, with excess generations trimmed where necessary to preserve $B=100$. On NPLIB1, the clearest improvement appears when the single-point query is replaced by a small grid of posterior queries; further refinement produces modest fluctuations without a consistent trend toward larger $M$. On MassSpecGym, performance improves more gradually before reaching a broad plateau around the default allocation. This pattern indicates that the main benefit comes from spanning plausible posterior operating points rather than repeatedly sampling from a single threshold. The default $M=20,N=5$ balances coverage across the query band with per-query sampling depth and lies within the stable region of both sweeps. \method{} maintains higher Top-10 accuracy than \methodbase{} throughout both sweeps, indicating that the advantage of posterior-aligned adaptation persists across candidate-pool allocations.

\begin{figure*}[t]
\centering
\includegraphics[width=\textwidth]{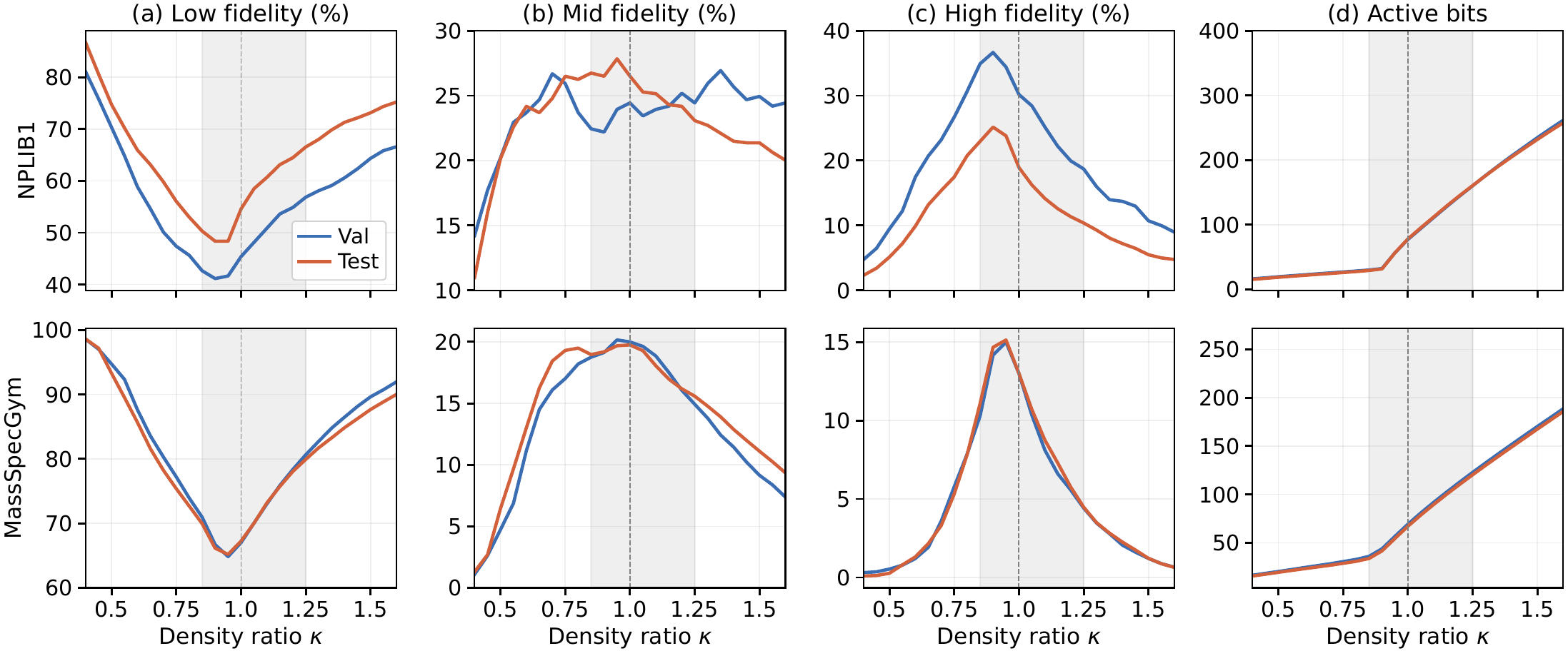}
\caption{Posterior-fidelity bin proportions (a--c) and mean active-bit count (d) versus density ratio $\kappa$ for NPLIB1 (top) and MassSpecGym (bottom). Curves compare validation and test splits; the dashed line marks density match ($\kappa{=}1.00$), and shading marks the calibrated density band $\mathcal{K}=[0.85,1.25]$.}
\Description{Two-row, four-column line plot with NPLIB1 in the top row and MassSpecGym in the bottom row. Columns (a), (b), and (c) show the proportions of low-fidelity, middle-fidelity, and high-fidelity spectra; column (d) shows mean active bits. Orange and blue lines denote test and validation splits respectively. The density ratio ranges from 0.4 to 1.6; a dashed line marks density ratio 1.0, and gray shading marks the calibrated density band from 0.85 to 1.25.}
\label{fig:kappa-diagnostic}
\end{figure*}

\subsection{Density-Ratio Diagnostics}
\label{sec:kappa-diagnostics}

Appendix~\ref{sec:band-config} specifies the shared endpoints $\kappa_{\min}$ and $\kappa_{\max}$, chosen empirically with reference to validation-set metrics. The resulting band spans both sides of density match, covering sparser queries as well as denser queries that retain a wider set of lower-confidence bits. This choice was guided by the expectation that co-occurrence patterns learned between active-bit conditioning tokens and SAFE tokens during conditional pretraining could help the model tolerate isolated noisy bits while benefiting from the additional structural cues retained by denser queries.

Figure~\ref{fig:kappa-diagnostic} provides a post hoc analysis of this calibration. Across the sweep, increasing $\kappa$ produces denser single-threshold queries, as confirmed by the monotonic active-bit curves. The high-fidelity share peaks slightly below density match; the middle-fidelity share remains elevated farther into the denser regime; and the low-fidelity share increases toward both extremes. To our surprise, validation and test exhibit similar $\kappa$-dependent patterns throughout the sweep. The curves nearly overlap on MassSpecGym; on NPLIB1, the absolute fidelity shares differ more, but the same qualitative progression is preserved. This cross-split agreement suggests that $\kappa$ provides a consistent coordinate for posterior fidelity across the observed splits. The high-fidelity peaks lie within the selected band, while its denser portion retains additional lower-confidence bits. These diagnostics illustrate the trade-off between posterior fidelity and the structural cues retained by the queries.

\subsection{Pretraining Data Scaling}
\label{sec:scaling}

Table~\ref{tab:scaling} examines a separate one-epoch pretraining trajectory of \methodbase{} on a filtered, approximately $1$B-structure SAFE-GPT corpus, with fingerprint--formula conditioning retained throughout. The architecture, candidate-pool size, and density-band inference protocol remain fixed, and connectivity-equivalent validation and test structures from both benchmarks are excluded. Its $100$M checkpoint is therefore distinct from the main two-epoch model. Top-1 accuracy and Morgan Tanimoto similarity improve throughout the sweep, indicating that greater pretraining exposure strengthens the leading prediction and structural similarity. Top-10 accuracy improves up to $100$M structures and then plateaus, with a small decline at the larger checkpoints. The gains therefore depend on the metric: further pretraining continues to improve Top-1 recovery and similarity, but does not yield corresponding gains in Top-10 accuracy.

\clearpage
\input{appendix_b_evaluation_alignment}

\clearpage
\section{Experimental Details}
\label{sec:exp-details}
\subsection{Model Architecture}

Our conditional molecule-language model augments the SAFE-GPT decoder~\cite{noutahi2024safe} with fingerprint and formula conditioning through cross-attention (Figure~\ref{fig:overview}(c)). The model contains $150.1$M unique parameters: an $86.7$M GPT-2 backbone warm-started from SAFE-GPT and $63.4$M parameters in the fingerprint encoder, formula encoder, cross-attention blocks, and shared context projection, all initialized from scratch. Table~\ref{tab:arch} summarizes the architecture and parameter counts.

\paragraph{Backbone.} The decoder is a $12$-layer GPT-2-style transformer with model width $768$, $12$ attention heads, and a maximum sequence length of $256$ tokens. Its vocabulary extends the $1880$-token SAFE-GPT vocabulary with the tokens \texttt{<} and \texttt{>}, giving $1882$ tokens in total. The language-model head shares its weight matrix with the input-token embedding. To warm-start the backbone, we copy the shared token-embedding rows, all transformer-block weights, the final layer norm, and the first $256$ positional-embedding rows from the public \texttt{datamol-io/safe-gpt} checkpoint; the two added token rows are initialized from scratch.

\paragraph{Fingerprint encoder.} We represent each query as a $4096$-bit radius-2 Morgan fingerprint~\cite{morgan1965algorithm,rogers2010ecfp} and retain only its active bits. Each selected bit is represented by a learned embedding of its index at width $768$. A $2$-layer pre-norm Transformer encoder ($12$ heads, feed-forward width $3072$, GELU) then contextualizes the per-bit embeddings. We retain at most $256$ active bits in ascending bit-index order and pad shorter sequences to this length with an accompanying validity mask. During conditional pretraining, the fingerprint input is the oracle fingerprint $\zstar$ computed from the molecule; at inference, it is a thresholded posterior query $z=\ind[\pi>t]$.

\paragraph{Formula encoder.} We represent the molecular formula as integer atom counts over $14$ elements (C, H, N, O, S, P, F, Cl, Br, I, B, Si, Se, As). For each element, a learned element embedding is concatenated with a two-layer MLP encoding of its atom count, projected to width $768$, and layer-normalized. This produces $14$ formula-conditioning tokens.

\paragraph{Cross-attention conditioning.} The padded fingerprint sequence and $14$ formula tokens are concatenated, together with their validity mask, into a shared context of $270$ token positions and passed through one shared linear projection. A conditioning block is inserted after decoder layers $2,4,6,8,10,$ and $12$. Each block applies pre-norm cross-attention from the decoder hidden states to the shared context, followed by a pre-norm feed-forward sublayer ($768\!\to\!3072\!\to\!768$, GELU); residual connections wrap both sublayers.

\begin{table*}[t]
\centering
\caption{Architecture and unique parameter counts of the conditional molecule-language model. The GPT-2 backbone is warm-started from SAFE-GPT, while the conditioning components are initialized from scratch. The tied token embedding and language-model head are counted once. Dropout is $0.1$ in the GPT-2 backbone, fingerprint Transformer, and cross-attention blocks.}
\label{tab:arch}
\setlength{\tabcolsep}{5pt}
\renewcommand{\arraystretch}{1.15}
\begin{tabular}{lll}
\toprule
Component & Configuration & \#Params \\
\midrule
Backbone (GPT-2)     & 12 layers, $d{=}768$, 12 heads, sequence length 256, vocabulary 1882 & $86.7$M \\
Fingerprint encoder  & 4096-bit radius-2 Morgan, $\le256$ active bits, 2-layer Transformer & $18.5$M \\
Formula encoder      & 14 elements, count MLP & $1.8$M \\
Cross-attention      & after layers 2, 4, 6, 8, 10, 12; 12 heads; FFN 3072 & $42.5$M \\
Context projection   & Linear $768\!\to\!768$ & $0.6$M \\
\midrule
Total                &                          & $150.1$M \\
\bottomrule
\end{tabular}
\end{table*}

\subsection{Conditional Pretraining}

Starting from the SAFE-GPT initialization described above, we train the fingerprint--formula-conditioned model with the molecule-only objective in Eq.~\eqref{eq:pretrain}; no spectra enter this stage. The conditionally pretrained decoder is shared across spectral datasets; within the decoder, only the lightweight adapter is trained separately for each dataset.

\paragraph{Corpus construction.} For the main model, we construct the corpus from a subset of the SAFE-GPT molecule collection. We convert SAFE strings to SMILES and remove stereochemistry before canonicalization, fingerprint computation, and formula extraction. We discard records that fail conversion, RDKit parsing, or feature computation, and exclude records whose 14-character InChIKey connectivity block matches that of a validation or test molecular structure in either benchmark. The resulting corpus contains $100{,}001{,}004$ records.

\paragraph{Formula-grouped batching.} We organize batches around exact molecular formula (Figure~\ref{fig:pretrain-adapt}(a)). Formula groups containing at least $64$ records, the per-GPU batch size, are shuffled internally and partitioned into complete same-formula batches. Smaller groups are pooled, shuffled, and partitioned into mixed-formula batches. Only complete local batches are retained; the combined batch list is shuffled each epoch and divided evenly among the eight data-parallel processes. This policy preserves the likelihood objective while increasing the within-batch concentration of alternatives that share the same formula.

\paragraph{Optimization.} We optimize all $150.1$M parameters for $2$ epochs with AdamW ($\beta=(0.9,0.95)$, weight decay $0.01$). The learning rate warms up for $2000$ updates to a peak of $3\times10^{-4}$ and then follows a cosine decay. Training employs bfloat16 mixed precision, gradient clipping at norm $1.0$, and a maximum sequence length of $256$. The effective global batch size is $1024$ (64 per GPU, eight GPUs, and two gradient-accumulation steps). The main two-epoch run requires approximately $358$ aggregate GPU-hours across eight NVIDIA A800 GPUs.

\subsection{Posterior-Aligned Adaptation}
\label{sec:adapt-details}

For each spectral dataset, we fit a separate posterior-aligned adapter to the shared conditional molecule-language model (Section~\ref{subsec:posterior-aligned-adaptation}, Figure~\ref{fig:pretrain-adapt}(b)).

\paragraph{Adapter placement.} We attach rank-$8$ LoRA modules ($\alpha=16$) for NPLIB1 and rank-$16$ modules ($\alpha=32$) for MassSpecGym, with dropout $0.05$, to $39$ linear layers: the fingerprint encoder's value and combine projections, the shared context projection, and six projections in each cross-attention block (query, key, value, output, and two feed-forward projections). All $150.1$M pretrained weights remain frozen; the $700{,}424$ and $1{,}400{,}848$ trainable LoRA parameters constitute $0.46\%$ and $0.92\%$ of the adapted models respectively. Before attaching LoRA to the query, key, and value projections, we split each fused query--key--value projection and copy the corresponding pretrained weight slices, leaving the model's initial function unchanged.

\begin{table}[t]
\centering
\caption{Recoverability groups and per-example loss weights for posterior-aligned adaptation (Eq.~\eqref{eq:pa_lora}).}
\label{tab:adapt}
\setlength{\tabcolsep}{5pt}
\renewcommand{\arraystretch}{1.1}
\begin{tabular}{llcc}
\toprule
Group & Criterion & CE $\alpha_g$ & KL $\lambda_g$ \\
\midrule
frozen-hit    & target in frozen $B{=}100$ pool & $0.0$ & $1.0$ \\
residual      & miss and $c\ge 0.40$        & $2.0$  & $0.5$ \\
low-coverage  & miss and $c<0.40$           & $0.1$  & $0.25$ \\
\bottomrule
\end{tabular}
\end{table}

\paragraph{Recoverability grouping.} The oracle fingerprint available for each annotated training pair serves only to assign its recoverability group. Under the default $B{=}100$ posterior-querying protocol, a pair is \textbf{frozen-hit} if the frozen candidate pool contains an InChIKey-14 match.

For a miss, we compute the posterior--oracle agreement
\(
c=\max_{t\in\mathcal{G}}\operatorname{Tanimoto}(\ind[\pi>t],\zstar),
\)
over the same dataset-specific $20$-point density-band grid $\mathcal{G}$ applied at inference (Appendix~\ref{sec:band-config}). Misses with $c\ge0.40$ are \textbf{residual}; the remainder are \textbf{low-coverage}. This grouping applies only during adaptation. The CE weight is largest for residual cases, small for low-coverage cases, and zero for frozen hits; the KL weight is nonzero for all three groups and largest for frozen hits (Table~\ref{tab:adapt}).

\paragraph{Objective and optimization.} For each example, we draw $t$ uniformly from its dataset-specific density-band interval and form $z_t=\ind[\pi>t]$. Equation~\eqref{eq:pa_lora} combines group-weighted teacher-forced SAFE cross-entropy with a forward-KL anchor from the frozen teacher to the adapted student; both receive the same $z_t$, and the KL is averaged over non-padding sequence positions. We optimize the LoRA parameters with AdamW ($\beta=(0.9,0.95)$, weight decay $0.01$), learning rate $10^{-4}$, a $6000$-step schedule with $150$ warmup steps followed by cosine decay, bfloat16 precision, gradient clipping at norm $1.0$, and maximum sequence length $256$. The global batch size is $64$ (32 examples on one GPU and two gradient-accumulation steps). We use checkpoints at steps $1000$ and $3000$ for NPLIB1 and MassSpecGym respectively.

\subsection{Density-Band Configuration}
\label{sec:band-config}

We apply a shared active-bit density-ratio band $\mathcal{K}=[0.85,1.25]$ to both benchmarks. The endpoints were chosen empirically with reference to validation-set metrics. The interval spans both sides of density match ($\kappa=1$), covering lower and higher active-bit densities relative to the oracle-fingerprint reference. We share the band in density-ratio space while deriving the corresponding thresholds from each encoder's training split.

We estimate $D_{\mathrm{enc}}$ and $A(t)$ from the corresponding encoder training split alone, obtaining $D_{\mathrm{enc}}=47.07$ active bits for NPLIB1 and $42.27$ for MassSpecGym. Inverting the endpoints of $\mathcal{K}$ through Eq.~\eqref{eq:opband} gives $(t_{1.25},t_{0.85})=(0.000943,0.9553)$ and $(0.03113,0.7947)$ respectively. At inference, we linearly interpolate $M{=}20$ thresholds between these endpoints to form $20$ query groups and draw $N{=}5$ sequences per group. Thus, the default candidate-pool size $B=M\times N=100$ denotes $100$ decoder samples per spectrum before postprocessing.

\subsection{Inference and Compute Benchmarking}
\label{sec:inference-details}

\paragraph{Posterior and generation.} For each benchmark, the corresponding dataset-specific MIST~\cite{goldman2023mist} checkpoint from the official release produces $\pi$ and remains frozen throughout our experiments. Given this posterior, \method{} constructs the dataset-specific threshold grid and samples SAFE sequences with temperature $1.0$, top-$k$ $50$, and top-$p$ $0.95$.

\paragraph{Postprocessing and evaluation.} We decode the generated SAFE sequences, canonicalize tautomers, remove stereochemistry, and group valid molecular structures by InChIKey-14. Posterior similarity is the maximum Morgan-fingerprint Tanimoto similarity between a candidate and the queries on the calibrated grid. The groups are ranked by formula matching, generation frequency and posterior similarity, in that order, and up to $k$ formula-matched structures are retained. When fewer than $k$ are available, we apply local atom and bond edits to candidates with mismatched formulas to obtain additional structures that match the input formula. Each parent contributes at most one new structure, selected by posterior similarity. Repaired structures inherit their parents' frequencies and are appended after the original formula matches, ordered by frequency and posterior similarity. All metrics are evaluated on the final ordered candidate list, with an output limit of $k=10$ for Top-1/Top-10 evaluation and $k=B$ for recall. We compute MCES with threshold $15$, stronger-bound selection enabled, and a $600$-second solver limit using CBC~\cite{liu2026massspecgymwild}.

\paragraph{Compute benchmarking.} We report per-spectrum inference compute for an NVIDIA RTX 6000 Ada GPU with a 16-core Intel CPU, including generation and Top-10 postprocessing for \method{}. To trace \method{} across candidate-pool sizes from $100$ to $1000$, we keep the $20$ query groups fixed and increase the number of samples per group. We measure MS-BART, MIST+MolForge, and MADGEN directly after warmup; the FRIGID and DiffMS timings are taken from FRIGID~\cite{bohde2026frigid}, which reports measurements on the same GPU.

\clearpage
\input{appendix_c_prediction_case_studies}

\end{document}

%% file: appendix_b_evaluation_alignment.tex
\section{Aligned Evaluations}
\label{sec:evaluation-alignment}

We further evaluate \method{} and \methodbase{} with predicted molecular formulas and a molecule-language model pretrained from scratch. We also archive corrected evaluations under the FRIGID MIST checkpoint setting used in our arXiv v1. Except for the conditions examined in each subsection, all experiments follow the main experimental setup, including the model implementation, hyperparameters, generation budget, postprocessing, and evaluation described in Appendix~\ref{sec:exp-details}. Paired table values report Top-1/Top-10.

\subsection{Unknown Molecular Formula Setting}
\label{sec:alignment-unknown-formula}

We evaluate \method{} and \methodbase{} on MassSpecGym with molecular formulas predicted by MIST-CF~\cite{goldman2024mistcf}, using the checkpoint released by MassSpecGym.\footnote{MIST-CF checkpoint: \href{https://huggingface.co/datasets/roman-bushuiev/MassSpecGym/blob/main/models/mist_cf_msg.ckpt}{\texttt{mist\_cf\_msg.ckpt}} on Hugging Face.} MIST-CF achieves Top-1 and Top-5 formula accuracies of $39.49\%$ and $64.32\%$, respectively. We allocate the fixed generation budget across the predicted formulas: the Top-1 setting uses 20 posterior queries with five samples per query, whereas the Top-5 setting uses four queries per formula with five samples per query. Each formula hypothesis is used for fragment annotation, fingerprint prediction, molecular generation, and postprocessing.

Table~\ref{tab:alignment-unknown-formula} shows that considering multiple formula hypotheses improves exact-match recovery for both models under the same generation budget. Although each formula receives fewer samples, the broader formula coverage increases the chance of recovering the target structure. Posterior-aligned adaptation improves Top-1 and Top-10 accuracy in both settings, indicating that its benefit extends to predicted formulas. Recovery nevertheless remains below the known-formula results, identifying formula prediction as an additional bottleneck in de novo structure elucidation from MS/MS spectra without a known molecular formula.

\begin{table}[htbp]
\centering
\caption{MassSpecGym results with MIST-CF-predicted formulas.}
\label{tab:alignment-unknown-formula}
\small
\setlength{\tabcolsep}{2pt}
\renewcommand{\arraystretch}{1.06}
\begin{tabular*}{\linewidth}{@{\extracolsep{\fill}}llccc@{}}
\toprule
\shortstack{Formula\\input} & Model & Accuracy\,$\uparrow$ & MCES\,$\downarrow$ & Tanimoto\,$\uparrow$ \\
\midrule
Top-1 & \methodbase{} & 5.18/6.57 & 29.71/28.20 & 0.24/0.27 \\
 & \method{} & \textbf{5.38}/\textbf{7.01} & \textbf{29.09}/\textbf{27.52} & \textbf{0.24}/\textbf{0.28} \\
\midrule
Top-5 & \methodbase{} & 6.37/8.04 & 29.96/23.16 & 0.24/0.32 \\
 & \method{} & \textbf{6.83}/\textbf{8.58} & \textbf{29.34}/\textbf{22.78} & \textbf{0.25}/\textbf{0.32} \\
\bottomrule
\end{tabular*}
\end{table}

\subsection{MLM Pretraining from Scratch}
\label{sec:alignment-50m-decoder}

We compare two pretraining routes for \methodbase{} and \method{}. In the first, the molecular language model is trained from random initialization for two epochs on the official MassSpecGym 50M-molecule corpus, followed by two epochs of conditional pretraining with fingerprints and formulas on the same corpus.\footnote{Pretraining corpus: \href{https://huggingface.co/datasets/roman-bushuiev/MassSpecGym/blob/main/data/molecules/MassSpecGym1.5_pretraining_molecules_50M_tani070.txt}{MassSpecGym1.5 50M molecules} on Hugging Face.} This corpus excludes molecules with Tanimoto similarity $\geq 0.7$ to any test molecule~\cite{liu2026massspecgymwild}. The main model starts from the official SAFE-GPT checkpoint and undergoes conditional pretraining on the 100M-molecule corpus. For each route, we evaluate the decoder before and after posterior-aligned adaptation on the MassSpecGym training split.

Table~\ref{tab:alignment-50m-decoder} shows that pretraining from scratch on the 50M corpus retains much of the main model's recovery performance despite the exclusion of close structural analogs. This suggests that transferable structural patterns learned during pretraining support recovery even without nearby pretraining molecules. Posterior-aligned adaptation improves exact-match accuracy with the main pretrained model, but slightly reduces it in the 50M setting despite improving MCES. This contrast suggests that a stronger pretrained molecular language model provides a better foundation for subsequent adaptation. LoRA adaptation also substantially improves adherence to the molecular-formula constraint in the 50M setting. Before postprocessing, the proportion of spectra with no formula-matched candidate decreases from $42.81\%$ for \methodbase{} to $27.98\%$ for \method{}. Postprocessing further reduces this proportion to below $1\%$ for all four models.

\begin{table}[htbp]
\centering
\caption{MassSpecGym results under two pretraining routes.}
\label{tab:alignment-50m-decoder}
\footnotesize
\setlength{\tabcolsep}{1.25pt}
\renewcommand{\arraystretch}{1.06}
\begin{tabular*}{\linewidth}{@{\extracolsep{\fill}}lllccc@{}}
\toprule
\begin{tabular}[c]{@{}c@{}}LM\\Pretraining\end{tabular} & \begin{tabular}[c]{@{}c@{}}Conditional\\Pretraining\end{tabular} & Model & Accuracy\,$\uparrow$ & MCES\,$\downarrow$ & Tanimoto\,$\uparrow$ \\
\midrule
50M corpus & 50M corpus & \methodbase{} & \textbf{8.88}/\textbf{11.55} & 13.19/11.73 & \textbf{0.38}/\textbf{0.43} \\
 &  & \method{} & 8.49/11.44 & \textbf{12.89}/\textbf{11.39} & 0.37/0.43 \\
\midrule
official ckpt & 100M corpus & \methodbase{} & 9.27/12.75 & 12.59/10.92 & 0.38/0.44 \\
 &  & \method{} & \textbf{10.22}/\textbf{13.21} & \textbf{12.09}/\textbf{10.55} & \textbf{0.39}/\textbf{0.45} \\
\bottomrule
\end{tabular*}
\end{table}

\subsection{Legacy FRIGID MIST Checkpoints}
\label{sec:alignment-legacy-mist}

We archive corrected evaluations of \methodbase{} and \method{} with the FRIGID MIST checkpoints used in our arXiv v1. The adapted models use adapters trained with the corresponding fingerprint posteriors. As reported by FRIGID, the legacy MassSpecGym checkpoint used synthetic training spectra generated from molecules selected for their structural similarity to test molecules, although exact test structures were excluded~\cite{bohde2026frigid}.

Table~\ref{tab:alignment-legacy-mist} includes the FRIGID arXiv v2 results reported in Table~12 of FRIGID~\cite{bohde2026frigid}. Under these checkpoint settings, posterior-aligned adaptation improves exact-match accuracy, MCES, and Top-10 Tanimoto similarity on both benchmarks.

\begin{table}[htbp]
\centering
\caption{Results with legacy FRIGID MIST checkpoints. MassSpecGym results cover 17,082 spectra for FRIGID and 17,556 for both \method{} variants.}
\label{tab:alignment-legacy-mist}
\small
\setlength{\tabcolsep}{2pt}
\renewcommand{\arraystretch}{1.06}
\begin{tabular*}{\linewidth}{@{\extracolsep{\fill}}lccc@{}}
\toprule
Model & Accuracy\,$\uparrow$ & MCES\,$\downarrow$ & Tanimoto\,$\uparrow$ \\
\midrule
\multicolumn{4}{c}{\textit{NPLIB1}} \\
\midrule
FRIGID (arXiv v2) & 25.03/33.37 & 7.10/5.56 & \textbf{0.58}/0.64 \\
\methodbase{} & 25.15/37.48 & 7.67/6.09 & 0.56/0.64 \\
\method{} & \textbf{26.62}/\textbf{42.98} & \textbf{6.49}/\textbf{4.90} & 0.57/\textbf{0.67} \\
\midrule
\multicolumn{4}{c}{\textit{MassSpecGym}} \\
\midrule
FRIGID (arXiv v2) & 18.29/22.00 & 13.49/11.65 & 0.43/0.47 \\
\methodbase{} & 23.66/31.94 & 9.33/7.64 & 0.54/0.61 \\
\method{} & \textbf{24.81}/\textbf{32.25} & \textbf{8.74}/\textbf{7.30} & \textbf{0.55}/\textbf{0.61} \\
\bottomrule
\end{tabular*}
\end{table}

%% file: appendix_c_prediction_case_studies.tex
% Case-study fragment. main.tex starts this appendix on a fresh page.

\section{Case Studies}
\label{sec:prediction-case-studies}

Under the fixed-grid inference and ranking protocol in
Appendix~\ref{sec:inference-details},
Figures~\ref{fig:prediction-cases-nplib1} and
\ref{fig:prediction-cases-msg} examine representative successes and failures
as candidate-pool size $B$ increases.  They show when candidate-pool scaling
changes Top-1 and distinguish finite-sample competition from frequency-dominated
misranking, missing candidate support, posterior mismatch, and fingerprint
resolution limits.

\subsection{Candidate-Pool Scaling and Candidate Recovery}

The stable rows in Figures~\ref{fig:prediction-cases-nplib1}(a) and
\ref{fig:prediction-cases-msg}(a) show that imperfect posterior evidence need
not preclude an exact Top-1 prediction.  With \emph{Fidelity} values of $0.849$
and $0.927$, respectively, both targets are Top-1 at $B{=}100$ and remain so
through $B{=}1000$.  More revealingly, the stable and recovery rows on NPLIB1
have nearly identical \emph{Fidelity} ($0.849$ and $0.850$), yet only the latter changes leader; fidelity
alone therefore does not determine sensitivity to candidate-pool size.

In the recovery rows, the decoys lead by only $2{:}1$ and $6{:}5$ at
$B{=}100$, while the targets have higher posterior similarity.  Both targets
become Top-1 at $B{=}200$ and remain there at $B{=}1000$.  These trajectories
are consistent with finite-sample near-ties: additional draws can overturn a
narrow frequency lead.  The cases therefore complement the aggregate result
in Appendix~\ref{sec:candidate-pool-scaling}: although larger pools primarily improve
candidate-pool recall, they can also change Top-1 when the target is already
present and the leading generation counts are close.

\subsection{Decoy Preference and Candidate Support}

The two NPLIB1 failures separate misranking from failure to generate the
target.  In Figure~\ref{fig:prediction-cases-nplib1}(c), the target remains
at rank~2 across all three budgets, and Top-1 never changes.  Although
the target is distinguishable from the decoy (\emph{Tani.} $=0.78$) and better
aligned with the posterior ($1.00$ versus $0.86$), the decoy leads $480{:}178$
at $B{=}1000$.  Thus, even with the target already present, the
decoy's persistent frequency advantage preserves the wrong leader as $B$ increases.

In Figure~\ref{fig:prediction-cases-nplib1}(d), the target remains absent
through $B{=}1000$.  The leading prediction replaces the seven-membered ring
with a six-membered ring plus a separate $[\mathrm{CH}_2]$, satisfying the
elemental composition but exposing the lack of a hard connectivity constraint
during token generation.

\subsection{Posterior Quality and Fingerprint Resolution}

\begin{figure*}[!t]
\centering
\includegraphics[width=\textwidth]{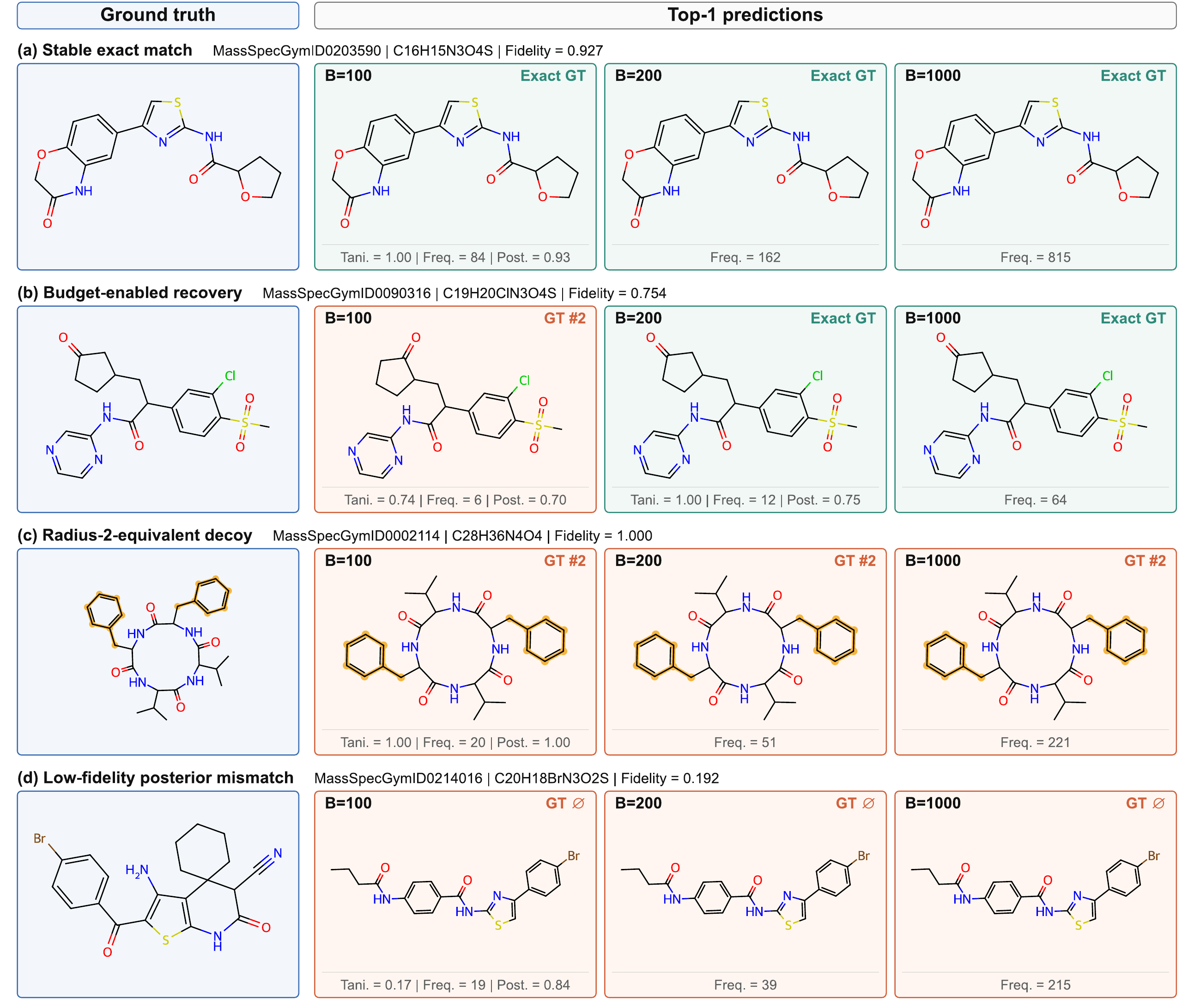}
\caption{MS-GPT case studies on MassSpecGym across candidate-pool sizes $B$.  Each
row compares the ground truth (blue) with Top-1 predictions (green: exact;
orange: decoy).  \emph{Fidelity} and \emph{Post.} are, respectively, the
maximum target--query and prediction--query radius-2 Morgan similarities over
the active-bit density band; \emph{Tani.} is the prediction--target radius-2 Morgan
similarity, and \emph{Freq.} is the grouped generation count.  For a repeated
Top-1 structure, only \emph{Freq.} is repeated because the other scores are
unchanged.  ``Exact GT,'' ``GT \#$k$,'' and ``GT $\varnothing$'' denote an
exact leading match, target rank, and target absence.  The gold highlight
marks the benzyl side chains.}
\Description{Four MassSpecGym examples show a stable exact match, recovery at
larger candidate-pool sizes, a failure caused by radius-2-equivalent target and
decoy structures, and an unrelated candidate strongly aligned with a
low-fidelity posterior.}
\label{fig:prediction-cases-msg}
\end{figure*}

Figure~\ref{fig:prediction-cases-msg}(d) exposes a low-fidelity posterior
mismatch.  \emph{Fidelity} is $0.192$, yet a dissimilar prediction
(\emph{Tani.} $=0.17$) reaches \emph{Post.} $=0.84$, while the target is never
generated.  The posterior is thus ``confidently wrong'' only operationally:
the query band matches the wrong fingerprint neighborhood, and increasing $B$
reinforces the resulting prediction but cannot alter the underlying posterior
mismatch.

\begin{figure*}[!t]
\centering
\includegraphics[width=\textwidth]{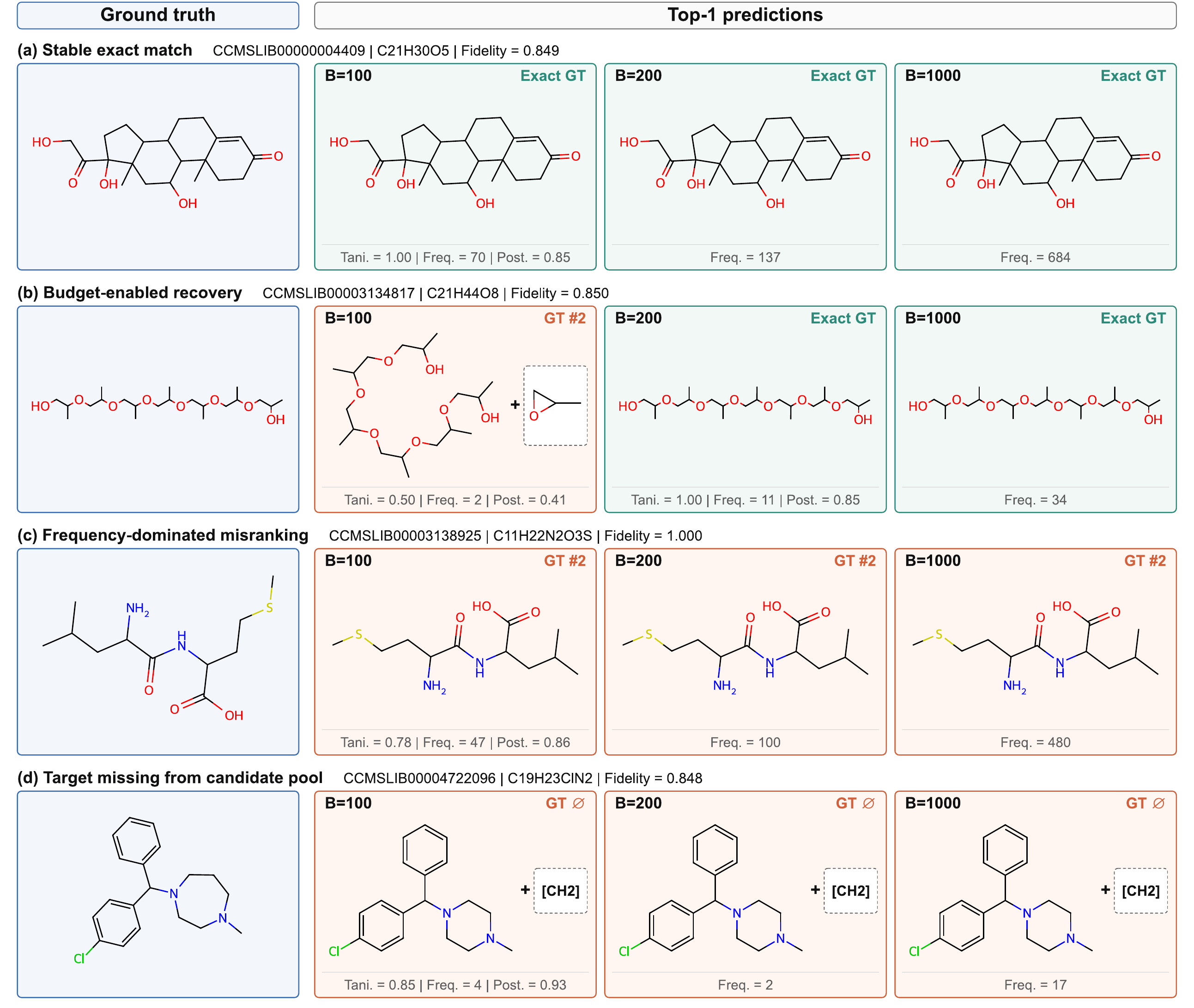}
\caption{MS-GPT case studies on NPLIB1 across candidate-pool sizes $B$.  Each row
compares the ground truth (blue) with Top-1 predictions (green: exact; orange:
decoy).  \emph{Fidelity} and \emph{Post.} are, respectively, the maximum
target--query and prediction--query radius-2 Morgan similarities over the
active-bit density band; \emph{Tani.} is the prediction--target radius-2 Morgan
similarity, and \emph{Freq.} is the grouped generation count.  For a repeated
Top-1 structure, only \emph{Freq.} is repeated because the other scores are
unchanged.  ``Exact GT,'' ``GT \#$k$,'' and ``GT $\varnothing$'' denote an
exact leading match, target rank, and target absence.  Dashed boxes mark
disconnected components, including the separate $[\mathrm{CH}_2]$ in (d).}
\Description{Four NPLIB1 examples compare the ground truth with MS-GPT Top-1
predictions at three candidate-pool sizes.  They show a stable exact match,
recovery at larger candidate-pool sizes, a persistent distinguishable decoy,
and a target absent from the candidate pool.}
\label{fig:prediction-cases-nplib1}
\end{figure*}

Figure~\ref{fig:prediction-cases-msg}(c) isolates a different failure: the
query band reaches the oracle radius-2 fingerprint, but that representation
itself is non-identifying.  \emph{Fidelity}, \emph{Post.}, and \emph{Tani.} are
all $1.00$, yet the target remains second at every candidate-pool size.  The target and
its cyclic-peptide sequence isomer have identical radius-2 Morgan fingerprints but
become distinguishable at radius~3.  Table~\ref{tab:radius-resolution}
contrasts this pair with the NPLIB1 case in
Figure~\ref{fig:prediction-cases-nplib1}(c), which is already distinguishable
at radius~2.

This non-identifiability persists even under the oracle radius-2 fingerprint,
making the failure orthogonal to the oracle-to-posterior gap and providing a
concrete representational limit of the fixed-dimensional fingerprint posterior
used by MS-GPT.  A higher-radius or multiscale condition can encode this
distinction, but whether the additional detail can be inferred reliably from
spectra remains open.

\begin{table}[H]
\centering
\caption{Target--decoy Morgan Tanimoto similarity across fingerprint radii for
Figures~\ref{fig:prediction-cases-msg}(c) and
\ref{fig:prediction-cases-nplib1}(c).  The MassSpecGym pair is
fingerprint-equivalent at radius~2 but becomes distinguishable at larger radii;
the NPLIB1 pair is already distinguishable at radius~2.}
\label{tab:radius-resolution}
\small
\setlength{\tabcolsep}{3.8pt}
\renewcommand{\arraystretch}{1.05}
\begin{tabular}{@{}llccc@{}}
\toprule
Spectrum & Structural change & $r{=}2$ & $r{=}3$ & $r{=}4$ \\
\midrule
\shortstack[l]{MassSpecGymID\\0002114} & Residue permutation & 1.00 & 0.91 & 0.77 \\
\midrule
\shortstack[l]{CCMSLIB\\00003138925} & Dipeptide reversal & 0.78 & 0.57 & 0.48 \\
\bottomrule
\end{tabular}
\end{table}

% Finish the analysis and Table~\ref{tab:radius-resolution} before the grids.
\clearpage

